\documentclass[10pt,twocolumn,letterpaper]{article}

 \usepackage[pagenumbers]{cvpr} 

\definecolor{cvprblue}{rgb}{0.21,0.49,0.74}
\usepackage[pagebackref,breaklinks,colorlinks,allcolors=cvprblue]{hyperref}
\usepackage{caption, float, multirow}
\usepackage{algorithm}
\usepackage{algpseudocode}
\usepackage{xcolor}
\usepackage{tcolorbox}
\usepackage{graphicx}
\usepackage{booktabs}
\usepackage{listings}
\algrenewcommand{\algorithmiccomment}[1]{\hfill\textcolor{green!50!black}{// #1}}

\def\paperID{12665} 
\def\confName{CVPR}
\def\confYear{2026}

\definecolor{codegreen}{rgb}{0,0.6,0}
\definecolor{codegray}{rgb}{0.5,0.5,0.5}
\definecolor{codepurple}{rgb}{0.58,0,0.82}
\definecolor{backcolour}{rgb}{0.95,0.95,0.92}

\lstdefinestyle{mystyle}{
	backgroundcolor=\color{backcolour},   
	commentstyle=\color{codegreen},
	keywordstyle=\color{magenta},
	numberstyle=\tiny\color{codegray},
	stringstyle=\color{codepurple},
	basicstyle=\ttfamily\scriptsize, 
	breakatwhitespace=false,         
	breaklines=true,                 
	captionpos=b,                    
	keepspaces=true,                 
	numbers=left,                    
	numbersep=5pt,                  
	showspaces=false,                
	showstringspaces=false,
	showtabs=false,                  
	tabsize=2
}
\title{InfiniBench: Infinite Benchmarking for Visual Spatial Reasoning with Customizable Scene Complexity}

\author{Haoming Wang, Qiyao Xue and Wei Gao\\
	University of Pittsburgh\\
	{\tt\small \{hw.wang, qiyao\_xue, weigao\}@pitt.edu}
}

\begin{document}
	
\makeatletter
\g@addto@macro\@maketitle{
	\begin{figure}[H]
		\setlength{\linewidth}{\textwidth}
		\setlength{\hsize}{\textwidth}
		\centering
		\vspace{-0.2in}
		\includegraphics[width=1\textwidth]{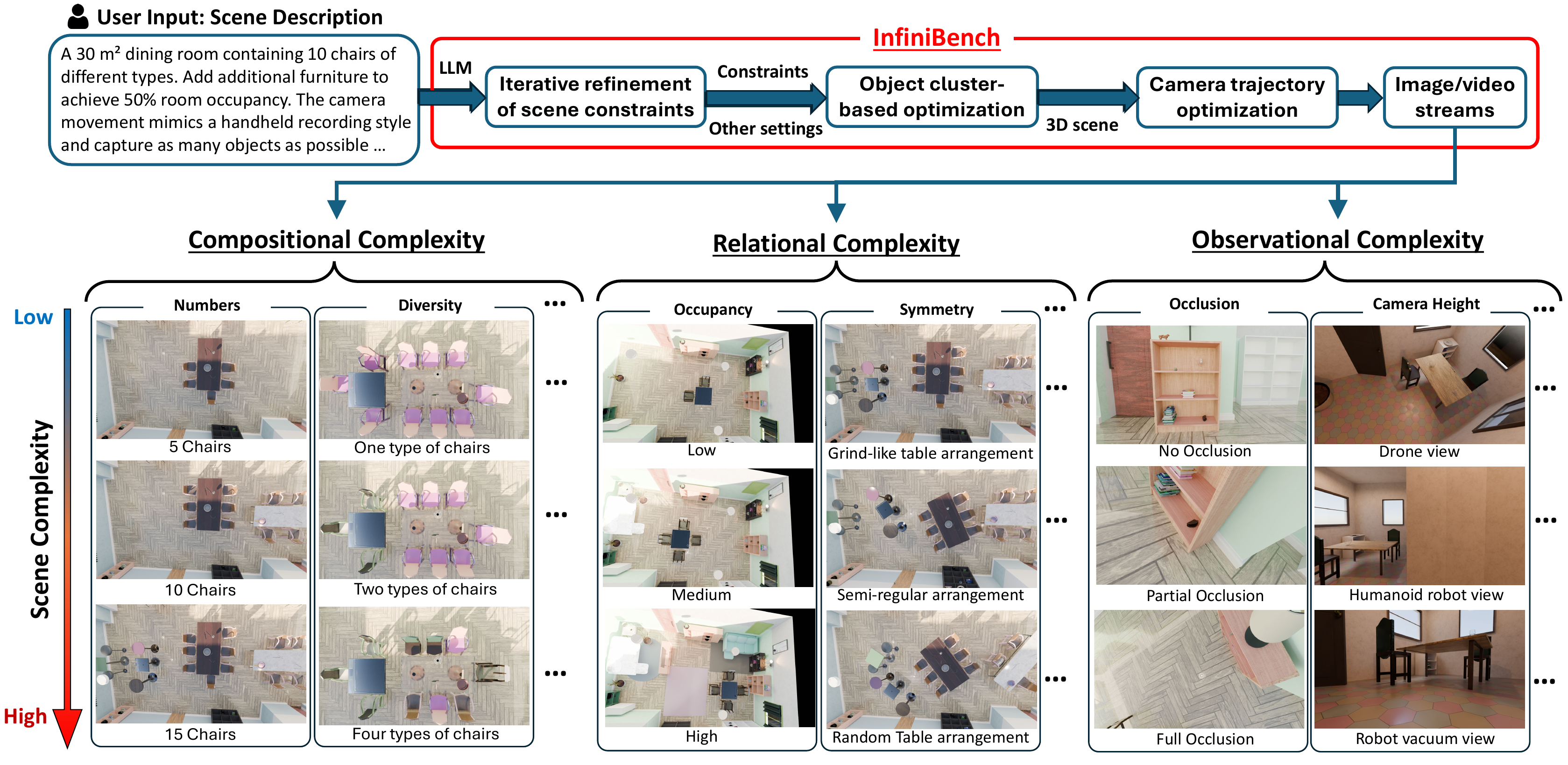}
		\vspace{-0.2in}
		\caption{\textbf{InfiniBench}, as a fully automated, customizable and user-friendly benchmark generator, can generate theoretically unlimited variations of 3D scene benchmarks, according to the parameterized constraints of compositional, relational and observational scene complexities, to meet different benchmarking needs of analyzing the VLM's successes and failures in spatial reasoning tasks.}
		\label{fig:overview}
		 \vspace{-0.05in}
	\end{figure}
}
\makeatother

\maketitle

\begin{abstract}

Modern vision-language models (VLMs) are expected to have abilities of spatial reasoning with diverse scene complexities, but evaluating such abilities is difficult due to the lack of benchmarks that are not only diverse and scalable but also fully customizable. Existing benchmarks offer limited customizability over the scene complexity and are incapable of isolating and analyzing specific VLM failure modes under distinct spatial conditions. To address this gap, instead of individually presenting benchmarks for different scene complexities, in this paper we present InfiniBench, a fully automated, customizable and user-friendly benchmark generator that can synthesize a theoretically infinite variety of 3D scenes with parameterized control on scene complexity. InfiniBench uniquely translates scene descriptions in natural language into photo-realistic videos with complex and physically plausible 3D layouts. This is achieved through three key innovations: 1) a LLM-based agentic framework that iteratively refines procedural scene constraints from scene descriptions; 2) a flexible cluster-based layout optimizer that generates dense and cluttered scenes previously intractable for procedural methods; and 3) a task-aware camera trajectory optimization method that renders scenes into videos with full object coverage as VLM input. Experiments demonstrate that InfiniBench outperforms state-of-the-art procedural and LLM-based 3D generation methods in prompt fidelity and physical plausibility, especially in high-complexity scenarios. We further showcased the usefulness of InfiniBench, by generating benchmarks for representative spatial reasoning tasks including measurement, perspective-taking and spatiotemporal tracking. The source codes of InfiniBench are available at \url{https://github.com/pittisl/infinibench}, and some sample benchmarks being generated by InfiniBench are available at \url{https://huggingface.co/datasets/Haoming645/infinibench}.
\end{abstract}

\vspace{-0.1in}
\section{Introduction}
Visual spatial reasoning \cite{yang2025thinking}, which comprehends the placement, orientation, and relationships of objects within a 3D space, is essential for AI to understand the real-world scenes from visual inputs such as images and video streams \cite{bai2023qwen,chen2024internvl,hurst2024gpt}. 
Unlike vision tasks on semantic perception of objects (e.g. image classification \cite{devadason2025comprehensive}, segmentation \cite{csurka2022semantic,barbosa2023threefoldreviewdeepsemantic} and captioning \cite{ghandi2023deep,hu2022scaling}), 
visual spatial reasoning requires abilities in geometric cognition, including spatial relation understanding, mental rotation and perspective taking \cite{stogiannidis2025mind}.

However, current vision-language models (VLMs) struggle to bridge the gap between 2D visual perception and 3D spatial reasoning, especially in scenes with high complexity \cite{chen2025gleam}. Such scene complexity is a multi-faceted concept that describes how the difficulty of understanding the scene's spatial arrangements varies, and it includes three dimensions: \emph{1) Compositional complexity} about the scene content such as the number and variety of objects, which could result in visual clutter \cite{chen2025gleam}; \emph{2) Relational complexity} about the intricacy of spatial relationships between objects, as more complicated spatial arrangement or higher ratio of scene occupancy challenges the VLM's capacity of accurately grounding the spatial language \cite{chen2025spatial}; \emph{3) Observational complexity} about how the scene is presented, encompassing challenges introduced by extreme viewpoints and object occlusion \cite{yu2025far}. These three dimensions disentangle different targets in human spatial perception, including the scene elements (compositional), the interconnections between these elements (relational), and the viewpoint-dependent nature of perception (observational) \cite{greene2023scene}.

To thoroughly understand the VLM's abilities of spatial reasoning in these dimensions is a big challenge, because \emph{they need benchmarks that are not only \textbf{diverse} and \textbf{scalable} but also \textbf{fully customizable}}. Benchmarks should enable full isolation and flexible manipulation of scene properties, such as object density, relative positioning and levels of occlusion levels, through a set of \textbf{configurable parameters} within a large combinatorial space. For example, to evaluate VLM's robustness to irrelevant objects, the benchmark should be able to systematically increase the number of irrelevant objects, while retaining the task-related objects unchanged. This fine-grained customizablility will allow us moving beyond the aggregate performance metrics (e.g., average task accuracy \cite{yang2025thinking}), to uncover the specific failure modes of VLMs under distinct spatial conditions.


Existing benchmarks face significant limitations in customizablility, scalability, and semantic richness. Real-world datasets \cite{yin2023lamm,hong20233d,jia2025omnispatial,wu2025spatial,chen2024spatialvlm,jia2025omnispatial}, although photorealistic, are costly to scale and lack parameterized control. A better alternative is to use synthetic data. Early methods use 3D rendering engines such as Blender \cite{blender} and IsaacSim \cite{isaacSim} to procedurally generate 3D scenes, but lack realism and fail to capture the real-world functional and aesthetic constraints \cite{wang20233d}. Recent techniques address these limitations by using 3D-aware diffusion models \cite{meng2025lt3sd,liu2024pyramid,go2025videorfsplat,go2025splatflow,engstler2025syncity,li2025worldgrow,xiang2025structured} with text \cite{yang2025prometheus,go2025videorfsplat} or image priors \cite{engstler2025syncity,xia2025scenepainter} for visual richness, but lack semantic annotations to generate QA pairs for reasoning tasks and fail to guarantee physical scene coherence. LLM-based scene generation \cite{ran2025direct,bucher2025respace,sun2025layoutvlm,zhang2025scene} can provide better semantic grounding, but falter on generating more complex scenes with correct layouts, due to LLMs' inherent limitations in spatial reasoning. Optimization-based procedural frameworks, such as Infinigen \cite{raistrick2024infinigen} and ProcTHOR \cite{deitke2022}, similarly fail to generate complex and cluttered scenes, and depend on professional tuning in practical use.

To address these challenges, we advocate a fundamentally different approach: instead of individually presenting benchmarks for different scene complexities, we present a fully automated, customizable and user-friendly \textbf{benchmark generator}, namely \textbf{\emph{InfiniBench}}, which can generate a theoretically unlimited number of 3D scenes from parameterized constraints of scene complexity given by users. The basic rationale of InfiniBench design is to combine the strengths of procedural 3D generation in scalability and control with the expressive power of natural language.
As shown in Figure \ref{fig:overview}, InfiniBench takes only a natural language description about the scenes as the input, and translates it into a physically plausible 3D scene that matches the scene complexity specified in the prompt. 

To achieve this, we first developed a new LLM-based agentic framework that wraps an advanced optimizer to generate and refine the 3D scene constraints, according to the given prompt of scene descriptions. The refined scene constraints are then transformed into 3D scenes by a custom optimization engine, which iteratively samples and places objects to satisfy all the semantic and physical rules in the scene constraints. In particular, we replace the conventional method of hierarchical optimization \cite{xu2002constraint,yu2011make,merrell2011interactive} with a more flexible cluster-based optimization strategy, to allow creating significantly more complex and dense scenes that were previously intractable. In this way, InfiniBench allows broad but effortless customization of the generated 3D scene from multiple perspectives (e.g., number and type of objects, spatial relationships, room layout, etc), without requiring professional knowledge or complicated tuning. 

Another gap between 3D scene generation and VLM benchmarking is that outputs of current scene generation methods, typically 3D models in Blender files or point clouds, are not directly consumable by VLMs. To generate image sequences or video streams that VLMs can properly perceive, we developed a new method of camera trajectory optimization, ensuring that all task-relevant objects and their spatial relationship are clearly visible and covered in camera views. Following the principle of frontier-based exploration \cite{yamauchi1998frontier,lluvia2021active,yamauchi1997frontier}, our algorithm incrementally builds the camera path by addressing each task-relevant object in sequence. For each object, it samples and evaluate multiple viewpoints and selects the best one that offers an unobstructed view after passing an occlusion check. A navigable path is then planned to link these viewpoints. 


We validate the efficacy of InfiniBench by generating scenes with different scene complexities and comparing the generated scenes with procedural and LLM-based generation methods. Results showed that InfiniBench simultaneously ensures prompt fidelity and physical plausibility, and successfully generates complex and high-density 3D scenes required by user prompts, where other methods fail. 

We further showed that customizable benchmarks generated by InfiniBench facilitate in-depth analysis of VLM's failure modes in various spatial reasoning tasks, including measurement (e.g. metric distance estimation), perspective-taking (e.g. object counting) and spatiotemporal tracking (e.g. identification of order identification) \cite{brown2025sims}. 
Such customizability in benchmarking would also be a key enabler of enhancing VLM's spatial reasoning capabilities, via spatially grounded training or prompt tuning.

\vspace{-0.05in}
\section{Related Work and Motivation}
\subsection{Variability of Scene Complexity}
\label{sec:controllable_complexity}
\vspace{-0.05in}

Existing spatial reasoning benchmarks \cite{chen2024spatialvlm,wu2025spatial,jia2025omnispatial} often categorize their data samples merely by task type or define scene complexity in simple terms, such as the number of rooms~\cite{chen2025gleam}. Such coarse-grained variability conflates multiple dimensions of the scene complexity, and hence obscures the precise reasons of VLM failures in reasoning tasks, which may include an excessive number of distractors (compositional complexity), an unusually intricate object arrangement (relational complexity), or heavily occluded viewpoints (observational complexity). Instead, in InfiniBench, we allowed much more fine-grained variability of scene complexity in these dimensions, by using separate sets of scene parameters (specified by users in natural language) to explicitly represent complexity levels of these dimensions and generating scenes accordingly.




\subsection{Layout Generation in 3D Scene Generation}


\begin{figure}
	\includegraphics[width=0.45\textwidth]{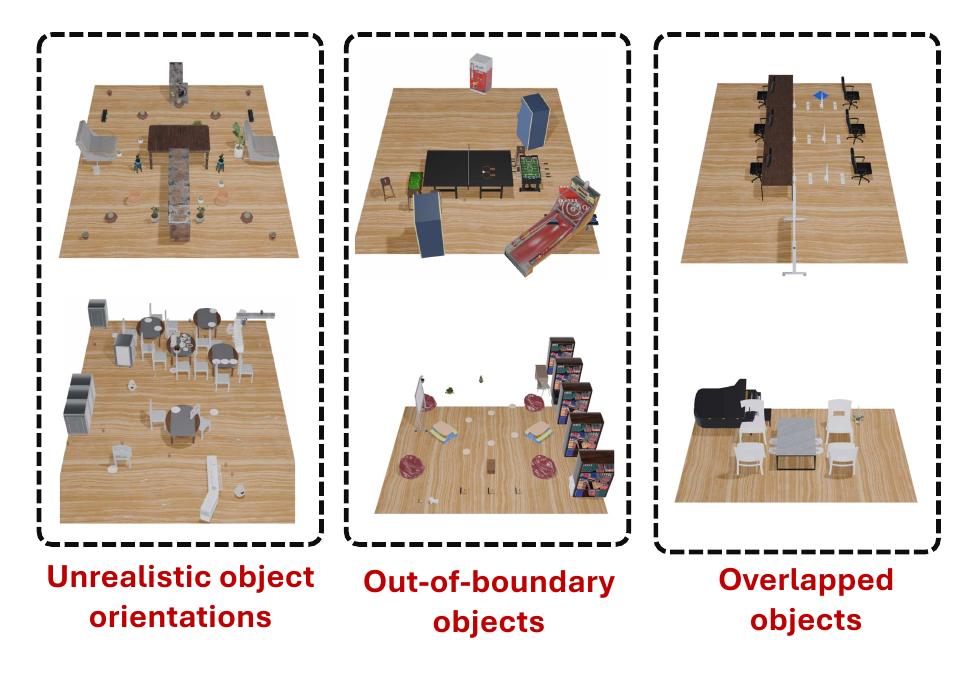}
	\caption{Limitations of LLM-based layout generation}
	\vspace{-0.1in}
	\label{fig:limitation_layout}
\end{figure}

The main focus of InfiniBench design is the creation of scene layout, which defines the placement, orientation, and relationships of objects in the scene.
One can leverage the reasoning capabilities of LLMs to directly produce a layout and then instruct a 3D engine to place assets in the scene \cite{ran2025direct,bucher2025respace,sun2025layoutvlm,zhang2025scene}. To improve the coherence of generated layouts, some works guide LLMs with structural templates \cite{feng2023layoutgpt} or represent scenes as graphs to facilitate compositional generation \cite{ccelen2024design,deng2025global,gao2024graphdreamer}. However, the generated scene's complexity is constrained by current LLMs' flawed spatial logic. As shown in Figure \ref{fig:limitation_layout},  as the number of objects and the intricacy of constraints increase, these methods frequently produce physically implausible layouts, such as those with unrealistic object orientations, out-of-boundary objects or overlapped objects.

To address this challenge and ensure the scalability of generating complex scenes, we proposed a more robust approach that separates high-level planning from low-level execution. Instead of tasking the LLM to generate the exact scene layout, we use LLM to generate high-level constraints about object arrangement. These constraints are then fed into a dedicated optimizer that generates a physically coherent 3D scene. By using optimization-based procedural generation frameworks \cite{raistrick2024infinigen,deitke2022} that are effective in 3D scene synthesis \cite{zhao2021luminous,yu2011make,yu2015clutterpalette}, we designed an improved optimizer for generating complex scenes and wrap this optimizer within an LLM-based agentic framework, making the entire pipeline both powerful and user-friendly.

\begin{figure}[ht]
	\vspace{-0.05in}
	\centering
	\includegraphics[width=\linewidth]{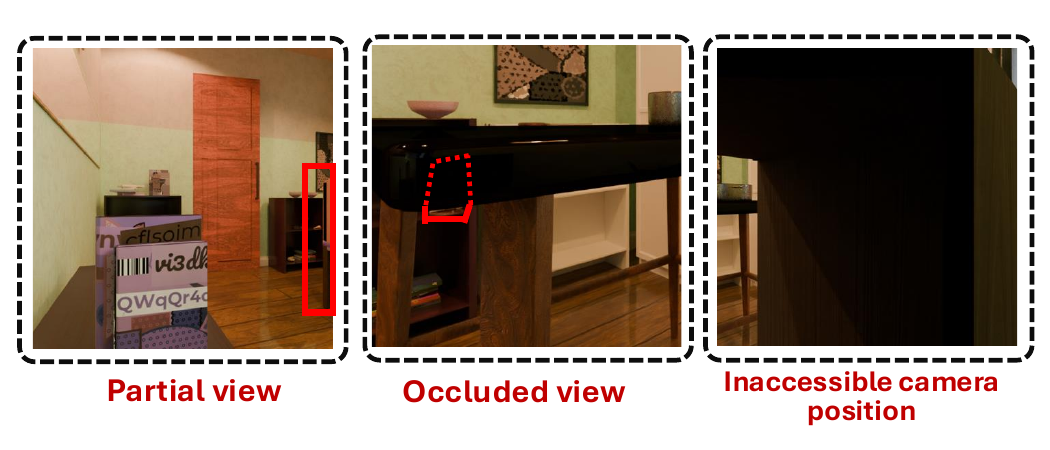}
	\vspace{-0.2in}
	\caption{Poorly chosen viewpoints obscure critical information}
	\label{fig:occlude_view}
\end{figure}

\subsection{Camera Trajectory Generation}
Once a 3D scene is generated, creating the 2D visual input for the VLM requires a carefully planned camera trajectory in the scene. The quality of this trajectory is paramount, because a poorly chosen viewpoint can easily obscure critical objects or relationships, rendering a spatial reasoning question unanswerable, as shown in Figure \ref{fig:occlude_view}. 

Optimization-based methods sample numerous camera parameters with respect to a predefined objective, to maximize the object visibility \cite{bares2000virtual,halper2001camera}. While offering fine-grained control, this approach is expensive and struggles with defining an objective function that robustly captures the visual requirements for complex spatial reasoning tasks. Another alternative is to train a generative AI model  \cite{courant2024exceptional,jiang2021camera,wang2024motionctrl}. However, existing datasets for such training are primarily for aesthetically pleasing and object-centric photo shots, e.g., smooth orbits around a single object, rather than trajectories with controlled coverage of multiple objects. Some other methods \cite{xie2023gait,dehghanian2025camera} used reinforcement learning to train an agent for scene navigation without large training data, but needs to retrain the agent for each new scene.

Instead, our approach is inspired by frontier-based exploration \cite{yamauchi1997frontier,yamauchi1998frontier,lluvia2021active}, a lightweight technique widely used in robotic navigation and mapping. In InfiniBench, we redefine frontier as object clusters, and sample viewpoints targeting these frontiers accordingly to strategically manage factors such as object occlusion. 


\begin{figure}
	\includegraphics[width=\linewidth]{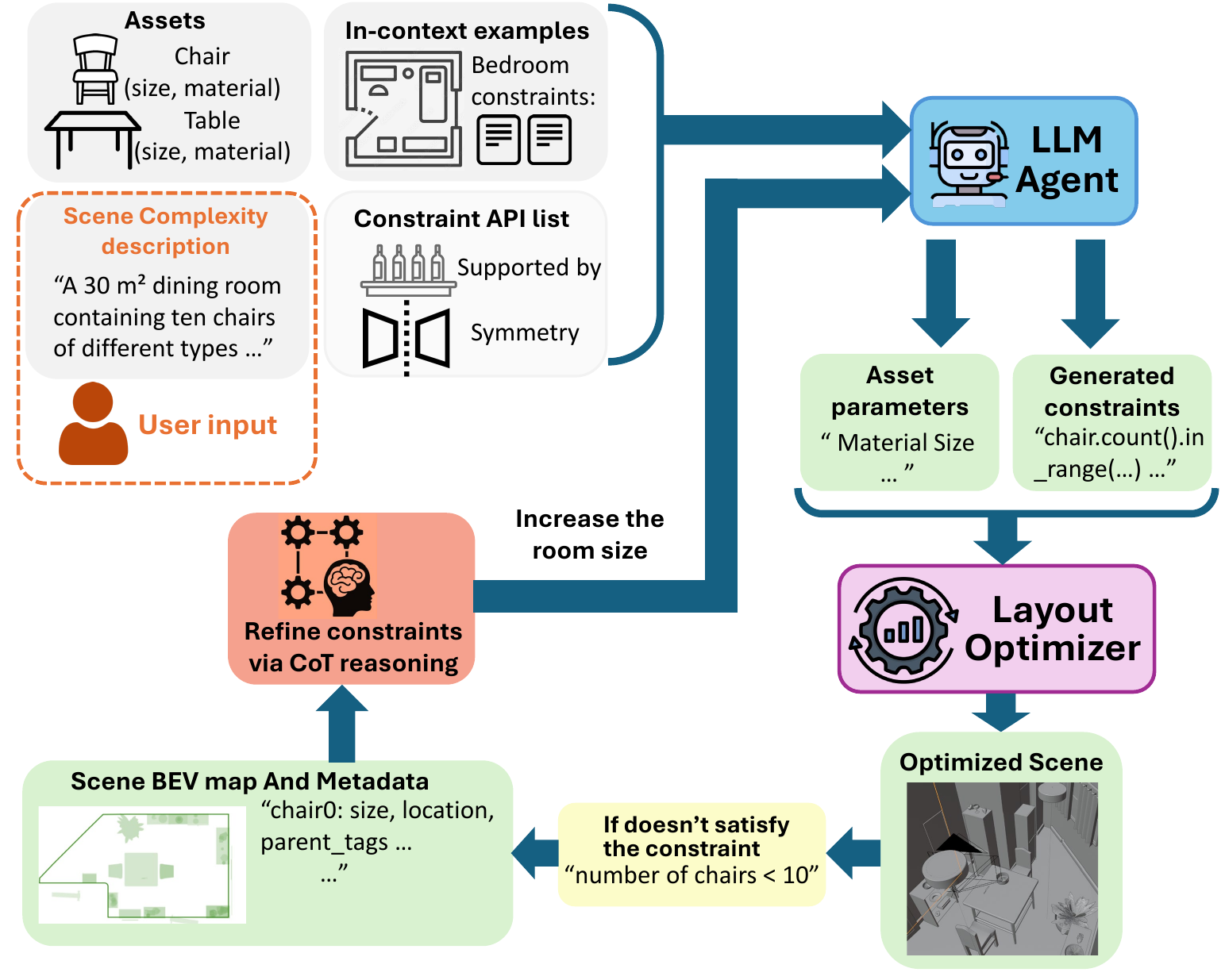}
	\vspace{-0.15in}
	\caption{LLM-based agentic framework for iterative scene constraint generation, illustrated by an example of generating a scene of a 30 m$^2$ dining room with 10 chairs of different types.} 
	\label{fig:constraint_generation}
		\vspace{-0.1in}
\end{figure}

\section{Method}
\label{sec:method}
As shown in Figure \ref{fig:overview}, the core of InfiniBench design is a three-stage pipeline, which consists of (1) an LLM-based agentic framework that translates user prompts of scene descriptions into procedural constraints through iterative refinement; (2) a cluster-based layout engine that optimizes these constraints into physically plausible 3D scenes; and (3) a task-aware camera trajectory algorithm that renders the 3D scene into an informative video as VLM input.

\subsection{Agentic Generation of Scene Constraints}
\label{subsec:agentic_constraint_generation}
A primary barrier to traditional procedural generation of 3D scenes is the need of manually scripting complex scene constraints, a process that requires significant technical expertise, especially when the scene complexity grows. InfiniBench democratizes this process, as shown in Figure \ref{fig:constraint_generation}, by using a LLM agent to translate scene descriptions in natural language into procedural constraints. To ensure that the generated constraints are machine-readable, we teach the LLM agent with domain knowledge of 1) a comprehensive list of available procedural APIs and 2) a set of few-shot examples. The information about procedural APIs covers detailed syntax and functionality with respect to the procedural 3D generation framework being used (e.g., Infinigen \cite{raistrick2024infinigen} and ProcTHOR \cite{deitke2022}). The few-shot examples demonstrate the translation from high-level user requests (e.g., ``a cluttered office desk'') to constraint definitions (e.g., \texttt{set\_object\_count(monitor, 3), on\_top\_of(keyboard, desk)}). 
More details about such in-context learning are in Appendix \ref{sec:appendix_a}.

A single pass of such scene constraint generation is often insufficient, as the LLM agent may generate constraints that are logically conflicting or physically impossible. For example, when asked to create a ``bedroom with three monitors'', it may generate constraints that place three monitors all on one desk asset with regular size, which is physically impossible due to the limited surface area of desk.

To address this challenge, InfiniBench iteratively refines the generated scene constraints via a critical feedback loop. In each iteration, we use the scene layout optimizer described in Section \ref{subsec:scene_layout_optimizer} to attempt realizing the generated constraints into a 3D scene. If the optimizer fails, the returned error report, which includes a bird's eye view (BEV) map showing object collisions and a textual summary describing the unmet constraints, is fed back to the LLM agent to refine the constraint generation. Such refinement enforces a Chain-of-Thought (CoT) reasoning process that prompts the LLM agent to first analyze the failure, propose a solution accordingly, and then implement the changes on scene constraints. For example, if the error report indicates that only one monitor is placed on the desk, the CoT reasoning process would infer the reason of failure as insufficient surface area, and the scene constraints will be accordingly revised to enforce larger parameters of the desk asset. More details about this CoT reasoning are in Appendix \ref{sec:appendix_b}. 

\begin{figure}
	\centering
	\includegraphics[width=0.42\textwidth]{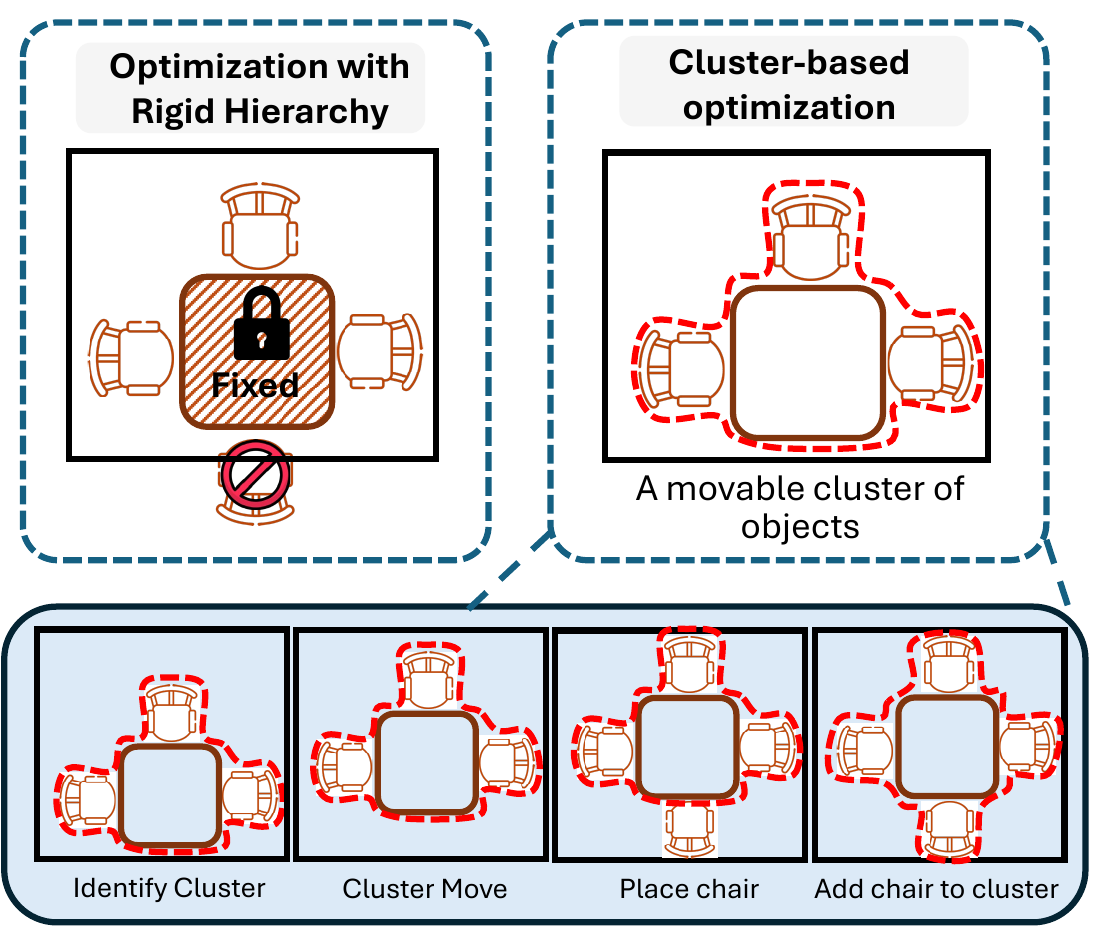}
	\caption{Traditional hierarchical optimization vs. our cluster-based optimization}
\label{fig:optimizer}
\vspace{-0.2in}
\end{figure}

\subsection{Layout Optimization for Complex Scenes}
\label{subsec:scene_layout_optimizer}
\begin{figure}[t]
	\centering
	
	\begin{subfigure}{0.45\textwidth}
		\centering
		\includegraphics[width=\linewidth]{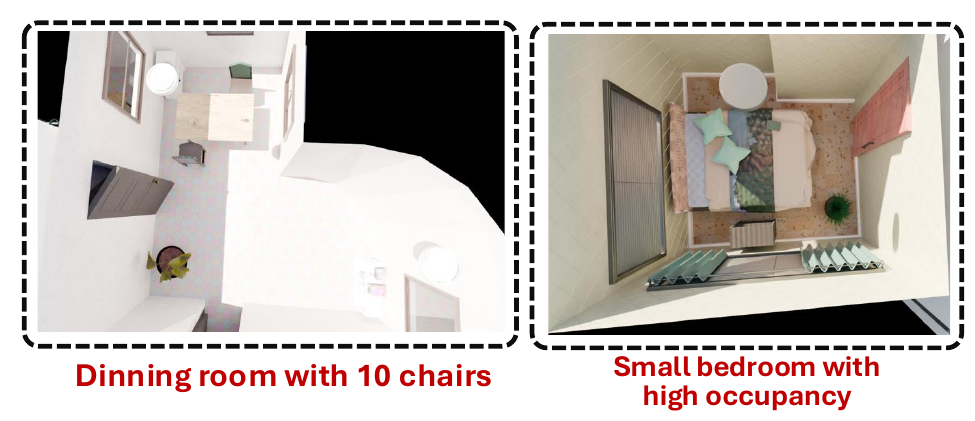}
		\caption{Scenes generated with traditional hierarchical optimization}
		\label{fig:subfig1}
	\end{subfigure}
	
	\begin{subfigure}{0.45\textwidth}
		\centering
		\includegraphics[width=\linewidth]{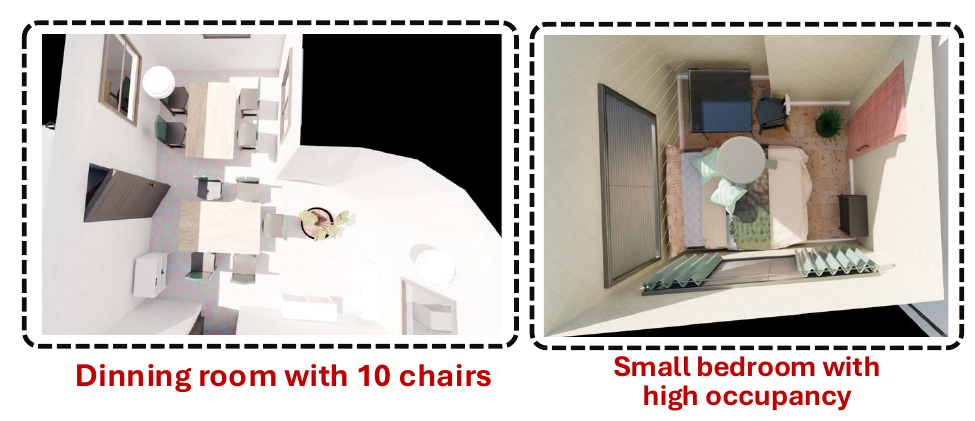}
		\caption{Scenes generated by our cluster-based optimization}
		\label{fig:subfig2}
	\end{subfigure}
	\vspace{-0.1in}
	\caption{Comparison of generating highly complex scenes, with the same room shape and asset parameters}
	\label{fig:optimizer_comparison}
\end{figure}

When the scene complexity grows, hierarchical optimization methods used in the existing procedural generation frameworks such as Infinigen \cite{raistrick2024infinigen} and ProcTHOR \cite{deitke2022} would fail, because they first fix the positions of large objects (e.g. desks and beds) and usually result in unsolvable states where there is no space for smaller objects (e.g. chairs around the desk), as shown in Figure \ref{fig:optimizer_comparison}(a). 

To ensure robust layout optimization with high scene complexity, we restructured the layout engine with a flexible cluster-based optimization strategy. This strategy, as shown in Figure \ref{fig:optimizer}, builds on a new concept of \textbf{movable cluster}, defined as a dynamic group of related objects (e.g., a table and chairs surrounding it) that are treated as a single entity in optimization. The optimization process, then, consists of the following three steps:
\begin{enumerate}
	\item \textbf{Identifying clusters:} The scene's semantic graph is parsed to automatically group closely related objects into clusters, each of which consists of a large ``parent'' object and a number of small ``child'' objects.
	\item \textbf{Augmenting the action space:} The optimizer's action space is expanded with cluster-level operations, which allow an entire cluster to move to a better location without breaking its internal relationships among objects.
	\item \textbf{Collision checking:} Collision checks are performed using each cluster's collective bounding box.
\end{enumerate}

This flexible and cluster-based approach allows the optimizer to navigate a much larger solution space and reposition the movable cluster for a valid layout, making the generation of previously intractable high-density scenes feasible, as shown in Figure \ref{fig:optimizer_comparison}(b). More details about such cluster operations are in Appendix \ref{sec:appendix_c}, and more examples of InfiniBench's generated scenes are in Appendix \ref{sec:appendix_d}.


\subsection{Camera Trajectory Optimization}

Given a generated 3D scene and a spatial reasoning task, the objective of our camera trajectory optimization in InfiniBench is to generate the shortest possible camera path that provides a clear, full and unobstructed view of every task-relevant object. Our method is inspired by frontier-based exploration \cite{yamauchi1997frontier,yamauchi1998frontier,lluvia2021active}, a classical robotic navigation technique. Unlike traditional approaches that explore frontiers using occupancy grids, we define ``frontiers'' as the set of unvisited target objects. We then develop a viewpoint sampling method to find optimal views for these targets while avoiding occlusion. As shown in Figure \ref{fig:path_gen}, this process consists of the following steps:

\begin{figure}
	\centering
	\includegraphics[width=0.95\linewidth]{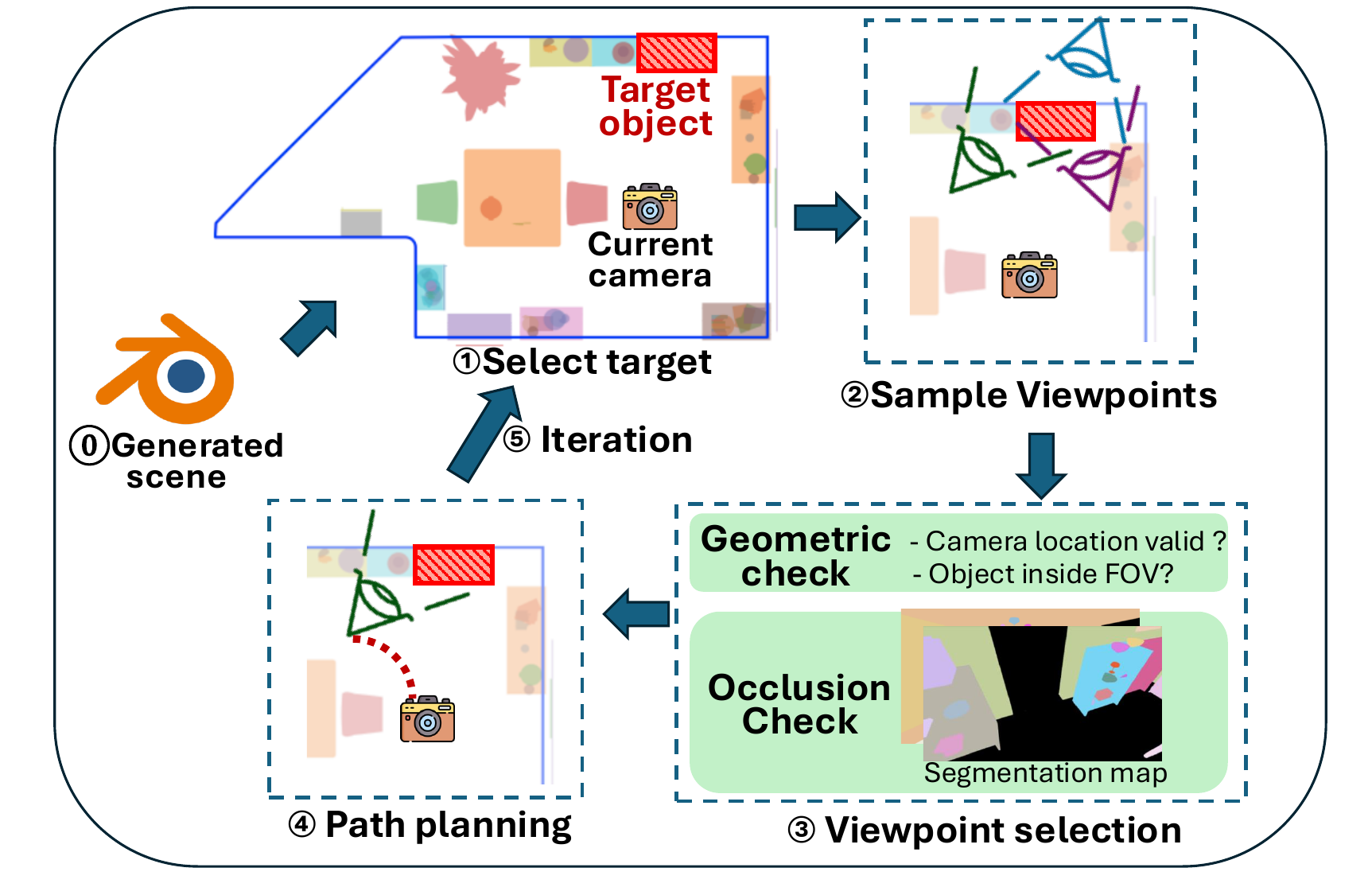}
	\caption{Camera trajectory optimization}
\label{fig:path_gen}
	\vspace{-0.1in}
\end{figure}

\begin{enumerate}
	\itemsep0em 
	\item \textbf{Target selection:} From the camera's current position, we identify the closest ``unvisited'' target object.
	\item \textbf{Viewpoint sampling and selection:} A set of candidate viewpoints is sampled around the target. Each viewpoint is scored based on three criteria: 1) if the camera location is valid, 2) if the object fully in the view? and 3) if there is occlusion between the camera location and the object. The viewpoint with the highest score is selected.
	\item \textbf{Path planning:} A collision-free path is computed from the current camera position to the selected viewpoint using Dijkstra's algorithm on a 2D floor plan.
	\item \textbf{Iteration:} The camera moves to the new viewpoint, the object is marked ``visited'', and the optimization repeats until all target objects are visited.
\end{enumerate}

After such optimization, we follow the optimized camera trajectory to render the scene into video frames, as exemplified in Figure \ref{fig:frames}. More details are in Appendix \ref{sec:appendix_e}.

\begin{figure}[ht]
	\vspace{-0.05in}
	\includegraphics[width=0.5\textwidth]{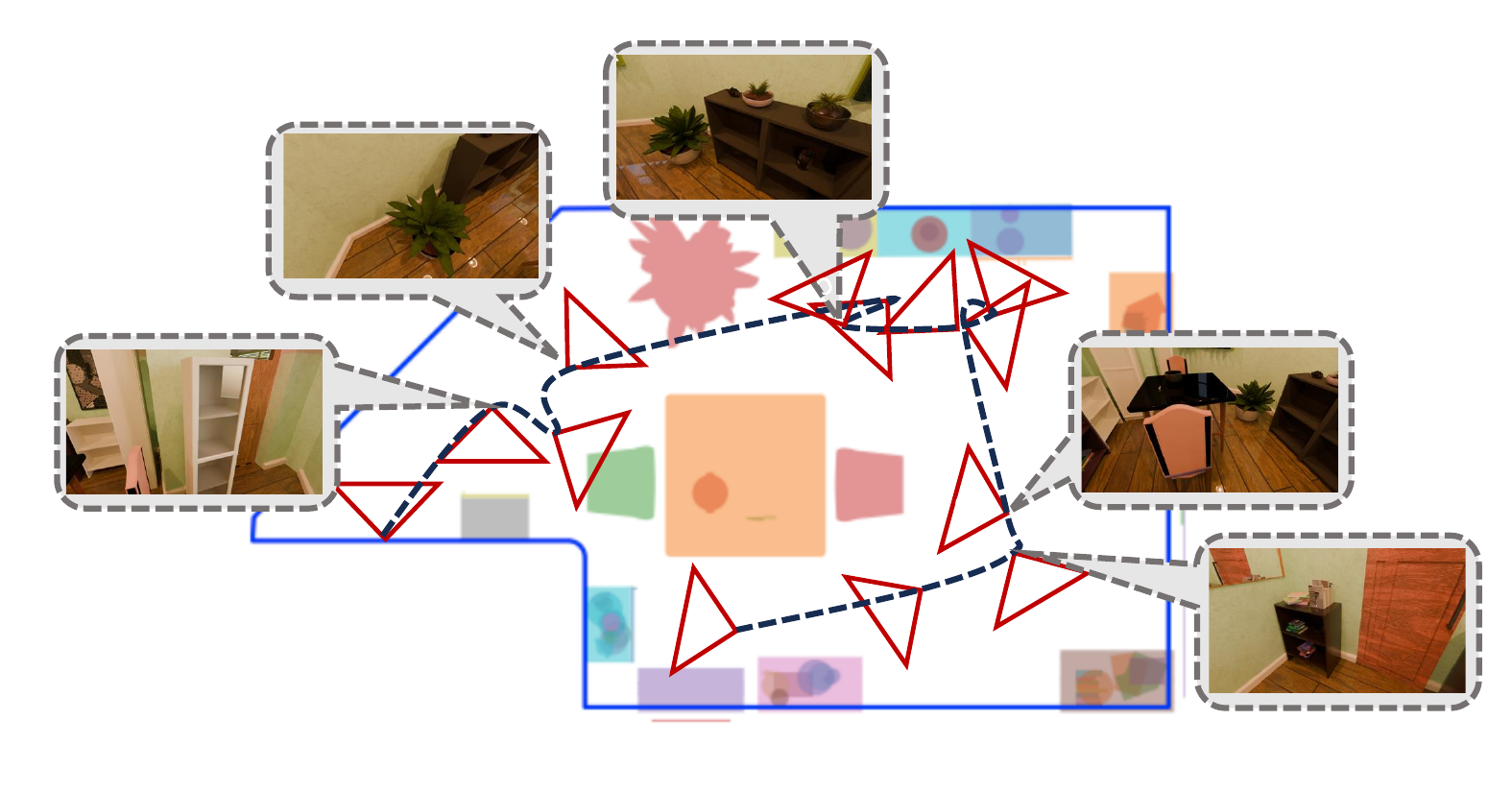}
	\vspace{-0.3in}
	\caption{An example of optimized camera trajectory and the corresponding video frames}
	\label{fig:frames}
	\vspace{-0.2in}
\end{figure}




\begin{table*}[t]
	\centering
	\resizebox{\textwidth}{!}{%
		\begin{tabular}{@{}lccccc|ccccc|ccccc@{}}
			\toprule
			& \multicolumn{5}{c}{\textbf{Low amount of objects}} & \multicolumn{5}{c}{\textbf{Medium amount of objects}} & \multicolumn{5}{c}{\textbf{High amount of objects}} \\
			\cmidrule(lr){2-6} \cmidrule(lr){7-11} \cmidrule(lr){12-16}
			\textbf{Method} & \textbf{Fidelity} $\uparrow$ & \textbf{CLIP} $\uparrow$ & \textbf{Realism} $\uparrow$ & \textbf{\#OB} $\downarrow$ & \textbf{\#CN} $\downarrow$ & \textbf{Fidelity} $\uparrow$ & \textbf{CLIP} $\uparrow$ & \textbf{Realism} $\uparrow$ & \textbf{\#OB} $\downarrow$ & \textbf{\#CN} $\downarrow$ & \textbf{Fidelity} $\uparrow$ & \textbf{CLIP} $\uparrow$ & \textbf{Realism} $\uparrow$ & \textbf{\#OB} $\downarrow$ & \textbf{\#CN} $\downarrow$ \\
			\midrule
			I-Design \cite{ccelen2024design} & \textbf{\underline{0.95}}& 28.4 & 0.72 & 1.6& 0.4& 0.92& 27.9 & 0.68 & 3.4& 5.1& 0.90& 27.1 & 0.61 & 6.9& 10.3\\
			Holodeck \cite{yang2024holodeck} & 0.97& 30.1 & 0.81 & 1.9& 2.1& 0.93& 29.5 & 0.77 & 4.6& 3.1& 0.88& 28.8 & 0.71 & 7.7& 9.4\\
			LayoutGPT \cite{feng2023layoutgpt} & 0.96& 29.8 & 0.79 & 1.3& 2.4& \textbf{\underline{0.95}} & 29.1 & 0.74 & 3.7& 4.6& 0.93 & 28.3 & 0.67 & 4.5& 13.5\\
			Luminous \cite{zhao2021luminous} & 0.87& 27.5 & 0.75 & \textbf{\underline{0.0}}& \textbf{\underline{0.0}}& 0.84& 26.8 & 0.70 & \textbf{\underline{0.0}}& 0.1& 0.42& 26.2 & 0.63 & \textbf{\underline{0.0}}& \textbf{\underline{0.0}}\\
			Infinigen \cite{raistrick2024infinigen} & 0.92& 31.0 & 0.88 & \textbf{\underline{0.0}} & 0.1& 0.87& 30.4 & 0.84 & \textbf{\underline{0.0}}& 0.2& 0.64& 29.7 & 0.79 & 0.0& 0.2\\
			\midrule
			\textbf{InfiniBench} & \textbf{\underline{0.98}}& \textbf{\underline{31.8}} & \textbf{\underline{0.93}} & \textbf{\underline{0.0}}& \textbf{\underline{0.0}}& \textbf{\underline{0.95}}& \textbf{\underline{31.5}}& \textbf{\underline{0.89}}& \textbf{\underline{0.0}}& \textbf{\underline{0.0}}& \textbf{\underline{0.98}}& \textbf{\underline{29.9}} & \textbf{\underline{0.81}} & 0.1& \textbf{\underline{0.0}}\\
			\bottomrule
		\end{tabular}%
	}
	\caption{Quantitative comparison of scene generation quality with different numbers of objects in the scene}
	\label{tab:object_results_revised}
\end{table*}

\begin{table*}[t]
	\centering
	\resizebox{\textwidth}{!}{%
		\begin{tabular}{@{}lccccc|ccccc|ccccc@{}}
			\toprule
			& \multicolumn{5}{c}{\textbf{Low Occupancy}} & \multicolumn{5}{c}{\textbf{Medium Occupancy}} & \multicolumn{5}{c}{\textbf{High Occupancy}} \\
			\cmidrule(lr){2-6} \cmidrule(lr){7-11} \cmidrule(lr){12-16}
			\textbf{Method} & \textbf{Fidelity} $\uparrow$ & \textbf{CLIP} $\uparrow$ & \textbf{Realism} $\uparrow$ & \textbf{\#OB} $\downarrow$ & \textbf{\#CN} $\downarrow$ & \textbf{Fidelity} $\uparrow$ & \textbf{CLIP} $\uparrow$ & \textbf{Realism} $\uparrow$ & \textbf{\#OB} $\downarrow$ & \textbf{\#CN} $\downarrow$ & \textbf{Fidelity} $\uparrow$ & \textbf{CLIP} $\uparrow$ & \textbf{Realism} $\uparrow$ & \textbf{\#OB} $\downarrow$ & \textbf{\#CN} $\downarrow$ \\
			\midrule
			I-Design \cite{ccelen2024design} & 0.90& 28.6 & 0.73 & 0.1& 0.4& 0.87& 28.0 & 0.67 & 1.3& 3.1& 0.79& 27.2 & 0.59 & 5.3& 6.4\\
			Holodeck \cite{yang2024holodeck} & 0.78& 30.4 & 0.82 & 0.3& 0.2& 0.80& 29.7 & 0.76 & 0.9& 5.4& 0.75& 28.9 & 0.69 & 3.1& 8.7\\
			LayoutGPT \cite{feng2023layoutgpt} & 0.81& 29.9 & 0.80 & 0.9& 0.2& 0.83& 29.3 & 0.73 & 1.7& 6.8& 0.85 & 28.5 & 0.65 & 5.6& 9.6\\
			Luminous \cite{zhao2021luminous} & 0.74& 27.8 & 0.76 & \underline{\textbf{0.0}}& \underline{\textbf{0.0}}& 0.64& 27.1 & 0.69 & \underline{\textbf{0.0}}& \underline{\textbf{0.0}}& 0.43& 26.4 & 0.61 & \underline{\textbf{0.0}}& \underline{\textbf{0.0}}\\
			Infinigen \cite{raistrick2024infinigen} & 0.83& 31.2 & 0.89 & \underline{\textbf{0.0}}& \underline{\textbf{0.0}}& 0.71& 30.6 & 0.85 & 0.2& 0.1& 0.49& 29.9 & 0.78 & 0.1& 0.2\\
			\midrule
			\textbf{InfiniBench} & \underline{\textbf{0.94}}& \underline{\textbf{32.1}} & \underline{\textbf{0.94}} & \underline{\textbf{0.0}}& \underline{\textbf{0.0}}& \underline{\textbf{0.89}}& \underline{\textbf{30.8}} & \underline{\textbf{0.86}} & \underline{\textbf{0.0}}& 0.1& \underline{\textbf{0.91}} & \underline{\textbf{30.2}} & \underline{\textbf{0.80}} & \underline{\textbf{0.0}} & 0.1\\
			\bottomrule
		\end{tabular}%
	}
	\caption{Quantitative comparison of scene generation quality with different levels of scene occupancy ratio}
	\label{tab:occupancy_results_revised}
\end{table*}

\section{Evaluation of InfiniBench}
\label{sec:experiment}

We evaluate InfiniBench to answer two primary questions: 1) Given the required scene complexity, can InfiniBench generate physically plausible 3D scenes that are superior to those from existing baseline methods, in terms of correctness and quality? 2) How effective are different components of InfiniBench in such generation? 

\subsection{Experimental Setup}
\label{sec:setup}

\noindent\textbf{Implementation.}
We use Gemini-2.5-Pro~\cite{comanici2025gemini} in the agentic framework for scene constraint generation, and our layout optimizer is built upon Infinigen~\cite{raistrick2024infinigen}, leveraging its rich and robust asset library. For camera trajectory generation and final 3D rendering, we enforce a pipeline with Trimesh~\cite{trimesh} for geometric processing, PyRender~\cite{pyrender} for fast occlusion checks, and Blender's Cycles engine~\cite{blender} for high-fidelity video output. More details are in Appendix \ref{sec:appendix_f}. 

\noindent\textbf{Baselines.}
We compare InfiniBench with two categories of 3D scene generation methods.
1) \emph{Procedural Frameworks} including Infinigen~\cite{raistrick2024infinigen} (using its original optimizer) and Luminous~\cite{zhao2021luminous}; and 
2) \emph{LLM-based Layout Generators} including LayoutGPT~\cite{feng2023layoutgpt}, Holodeck~\cite{yang2024holodeck} and I-Design~\cite{ccelen2024design}, which directly generate scene layouts from text. For fair comparisons, the same 3D asset library (provided by Infinigen) is used in all experiments.

\noindent\textbf{Evaluation Metrics.}
We focus on two challenging aspects of scene complexity: the total number of objects and the scene occupancy ratio. We evaluate the generated scenes using the following metrics:
\begin{itemize}
	\itemsep0em
	\item \emph{Prompt fidelity}, which measures the alignment between the generated scene and text prompt. It is calculated as the accuracy of the generated scene's object count and occupancy ratio against ground truth values in text prompt.
	\item \emph{Text-image alignment}, which measure the semantic alignment as the CLIP score ~\cite{radford2021learning} between the text prompt and the rendered top-down image of generated scene.
	\item \emph{Layout realism}, scored by a GPT-5 evaluator as suggested by Scenethesis~\cite{ling2025scenethesis}, incorporating key factors such as spatial coherence and functional plausibility.
	\item \emph{Physical plausibility}, measured by the number of physically impossible artifacts, especially the count of out-of-boundary objects (\#OB) and pairs of collided objects (\#CN), as suggested by SceneWeaver~\cite{yang2025sceneweaver}.
\end{itemize}

\begin{figure*}[ht]
	\centering
	\includegraphics[width=0.95\textwidth]{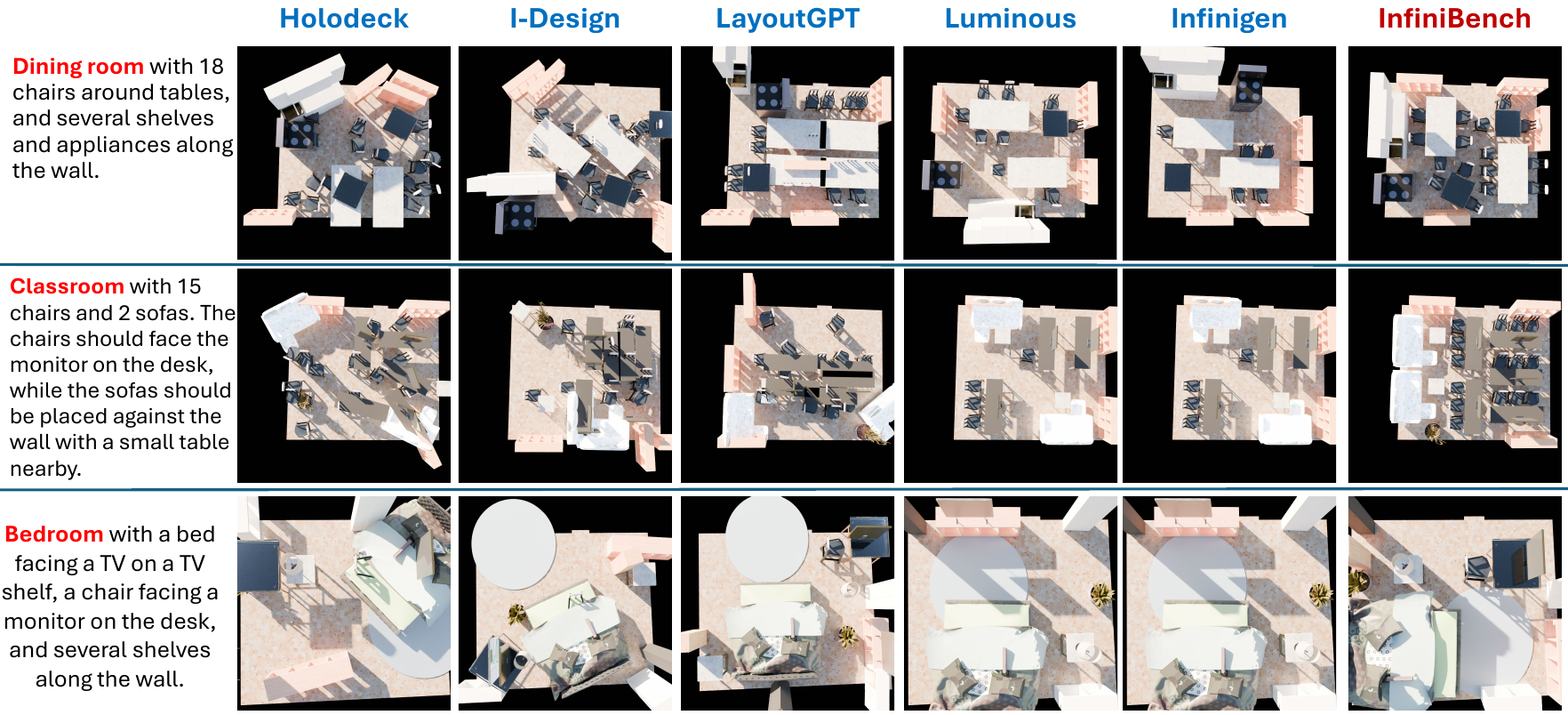}
	\caption{Qualitative comparisons of the scenes generated by InfiniBench and baseline methods}
	\vspace{-0.1in}
	\label{fig:qualitative_comparison}
\end{figure*}

\subsection{Main Results}

\noindent\textbf{Quantitative Results.}
Tables~\ref{tab:object_results_revised} and~\ref{tab:occupancy_results_revised} show that as the scene complexity increases, current generation methods explicit distinct tradeoffs among different evaluation metrics. LLM-based generation methods reach the required complexity constraints and maintain relatively high Fidelity at the cost of lower physical plausibility, as evidenced by an increase in \#OB and \#CN. For instance, with a high amount of objects in the scene, LayoutGPT's collision count rises to 13.5, 5.6x higher than that with a low amount of objects. Conversely, procedural generation frameworks excel at maintaining physical plausibility, but have difficulty in adhering to the prompt's quantitative requirements of scene complexity. As such required scene complexity increases, their Fidelity scores drop drastically (e.g., 0.42 and 0.64 for Luminous and Infinigen, respectively), indicating a failure of generating the required number of objects.

InfiniBench successfully avoids such tradeoff. It consistently achieves high Fidelity comparable to the LLM-based approaches, and simultaneously ensures near-perfect physical plausibility that is on par with procedural methods. In the most challenging high-complexity settings, InfiniBench reaches top performance in all metrics, including the highest CLIP and Realism scores, demonstrating its superiority of generating both complex and physically coherent scenes.

\noindent\textbf{Qualitative Results.}
Figure~\ref{fig:qualitative_comparison} provides qualitative comparisons of the scenes generated by InfiniBench and baseline methods. The scenes generated by baseline methods often fail in predictable ways when the required scene complexity high: the generated objects could be haphazardly placed, frequently intersecting, or fail to appear at all. In contrast, InfiniBench produces coherent, cluttered and physically plausible scenes that faithfully reflect the user's detailed request. This visual evidence underscores the effectiveness of our iterative refinement and cluster-based optimization pipeline. More examples are in Appendix \ref{sec:appendix_g}.


\noindent\textbf{Ablation Study.}
To validate the contribution of each component in InfiniBench, we conduct an ablation study. We start with the base optimizer provided in InfiniGen, and compare it against with \textit{constraint-refinement only}, \textit{cluster-optimization only}, and the full InfiniBench that includes both agentic framework for constraint refinement and cluster-based layout optimizer. Results in Table~\ref{tab:ablation} show that, even without the cluster-based layout optimizer, \textit{constraint-refinement only} provides a moderate boost to Fidelity (0.64 $\rightarrow$ 0.71), demonstrating its effectiveness in better following the text prompt. \textit{Cluster-optimization only} yields a slighter improvement in Fidelity (to 0.68) but also marginally increases Realism. Combining both components results in a dramatic and synergistic improvement.

\begin{table}[ht]
	\centering
	\small
	{\fontsize{8}{9}\selectfont
		\begin{tabular}{cccccc}
			\toprule
			& Fidelity & Clip & Realism & \#OB &\#CN\\
			\midrule
			Base & 0.64 & 29.7 & 0.79 & 0.0& 0.0\\ 
			Constraint-refinement only & 0.71 & 28.9 & 0.79 & 0.0& 0.0\\ 
			Cluster-optimization only & 0.68 & 29.9 & 0.81 & 0.0&0.1 \\ 
			Full InfiniBench & 0.92 & 29.9 & 0.81 & 0.2 & 0.1\\ 
			\bottomrule
	\end{tabular}}
	\caption{Ablation study of different components in InfiniBench}
	\label{tab:ablation}
\end{table}

Further, we also evaluated how iterative refinement of scene constraints (Section \ref{subsec:agentic_constraint_generation}) affects scene generation. As shown in Table \ref{tab:ablation_iteration}, the refinement process typically converges within 5 iterations. While our method does require iterative reasoning, the resulting constraints are highly reusable. The layout optimizer can leverage these constraints to generate a large number of scenes, hence ensuring the high compute efficiency in layout generation.

\begin{table}[ht]
	\centering
	\small
	{\fontsize{8}{9}\selectfont
	\begin{tabular}{cccccccc}
		\toprule
		& Fidelity & Clip & Realism & \#OB &\#CN \\
		\midrule
		1 iteration & 0.68& 29.9&0.81&0.1&0.2 \\
		2 iteration & 0.72& 29.8&0.82&0.1&0.1 \\
		3 iteration & 0.86& 29.9&0.80&0.0&0.2 \\
		5 iterations &0.92 &29.9 & 0.81&0.1&0.1\\
		10 iterations &0.92 &29.8 & 0.81&0.2&0.1\\
		\bottomrule		
	\end{tabular}}
	\caption{Ablation study of how the number of iterations in constraint refinement affects the scene quality}
	\label{tab:ablation_iteration}
\end{table}

\begin{table*}[ht]
	\centering
	\small
	{\fontsize{8}{9}\selectfont
		\begin{tabular}{l|ccc|ccc|ccc}
			\toprule
			\textbf{Amount of irrelevant objects} & \multicolumn{3}{c|}{\textbf{Measurement task}} & \multicolumn{3}{c|}{\textbf{Perspective-taking  task}} & \multicolumn{3}{c}{\textbf{Spatiotemporal task}} \\
			\cmidrule(lr){2-4} \cmidrule(lr){5-7} \cmidrule(lr){8-10}
			& \textbf{Low} & \textbf{Medium} & \textbf{High} & \textbf{Low} & \textbf{Medium} & \textbf{High} & \textbf{Low} & \textbf{Medium} & \textbf{High} \\
			\midrule
			Gemini-2.5-pro & 69.2 & 68.9 & 66.4 & 70.6 & 67.2 & 66.9 & 87.9 & 70.1 & 56.2\\
			GPT5 & 45.8 & 41.2 & 41.3 & 57.3 & 55.5 & 54.1 & 47.8 & 31.3 & 26.7\\
			LLaVA-Video-7B & 47.2 & 45.2 & 45.0 & 56.0 & 56.1 & 54.2 & 50.2 & 36.2 & 29.2\\
			InternVL3.5-8B & 48.4 & 48.2 & 47.8 & 53.0 & 52.9 & 52.1 & 79.8 & 64.9 & 47.0\\
			Cambrian-S-7B & 57.2 & 57.0 & 56.9 & 69.1 & 71.3 & 65.4 & 85.1 & 72.4 & 49.2\\
			\bottomrule
	\end{tabular}}
	\caption{Performance of Measurement, Perspective-taking and Spatiotemporal tasks with different numbers of irrelevant objects}
	\label{tab:combined_performance}
\end{table*}

\section{Using InfiniBench to Diagnose VLMs' Failures in Spatial Reasoning}
\label{sec:case_studies}
To further verify the usefulness of InfiniBench in diagnosing VLMs' abilities and failure modes in spatial reasoning with high complexity, we apply the benchmarks generated by InfiniBench with different levels of compositional, relational and observational complexities to the following set of representative spatial reasoning tasks \cite{brown2025sims}. We examine where and why VLM's spatial reasoning succeeds or fails, with two close-sourced models (Gemini-2.5-Pro \cite{comanici2025gemini}, GPT-5 \cite{openai_gpt5_system_card}) and three open-sourced models (LLaVa-Video-7B \cite{zhang2024videoinstructiontuningsynthetic}, InternVL3.5-8B \cite{wang2025internvl3_5}, Cambrian-S-7B \cite{yang2025cambrian}). 

\noindent\textbf{Measurement tasks} that probe VLM's ability of reasoning about metric properties and object scale. We generate tasks that require both dimensional comparison and fine-grained distance estimation based on complex referring expressions, such as \textit{Q: What's the [Height] of the [Object\_Name] with the texture of [Material\_name] and [Relation to a nearby object]}. Performance will be measured by mean relative accuracy between numerical answer and ground truth \cite{yang2025thinking}.

\noindent\textbf{Perspective-taking tasks} that assess compositional understanding from a given viewpoint. We choose object counting tasks with various numbers of irrelevant objects, such as \textit{Q: How many [Object\_Name] are in the video}. Performance will be similarly measured by mean relative accuracy \cite{yang2025thinking}.

\noindent\textbf{Spatiotemporal tasks} that test the VLM's ability of reasoning over time and space. We generate tasks of identifying the appearance order of objects that demand temporal memory and ability of tracking objects across a dynamic camera trajectory, such as \textit{Q: What's the appearance order of [Obj\_A\_description], [Obj\_B\_description], and [Obj\_C\_description]}. Tasks are multi-choice questions, and performance is measured by exact match accuracy.

\begin{figure}[ht]
	\vspace{-0.05in}
	\includegraphics[width=\linewidth]{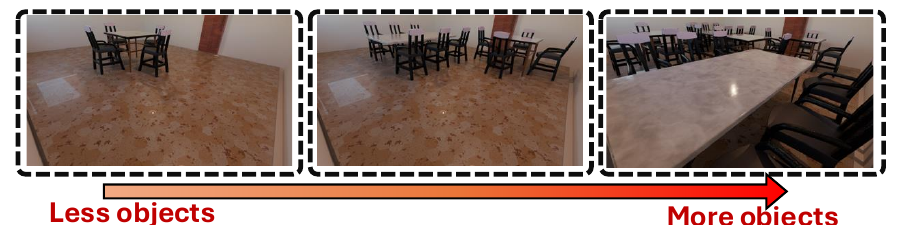}
	\vspace{-0.1in}
	\caption{Examples of benchmark samples with varying compositional complexity (i.e., the number of objects)}
	\label{fig:number}
\end{figure}

\subsection{VLM's Spatial Reasoning with Varying Compositional Scene Complexity}
\begin{figure}
	\centering
	\includegraphics[width=0.4\textwidth]{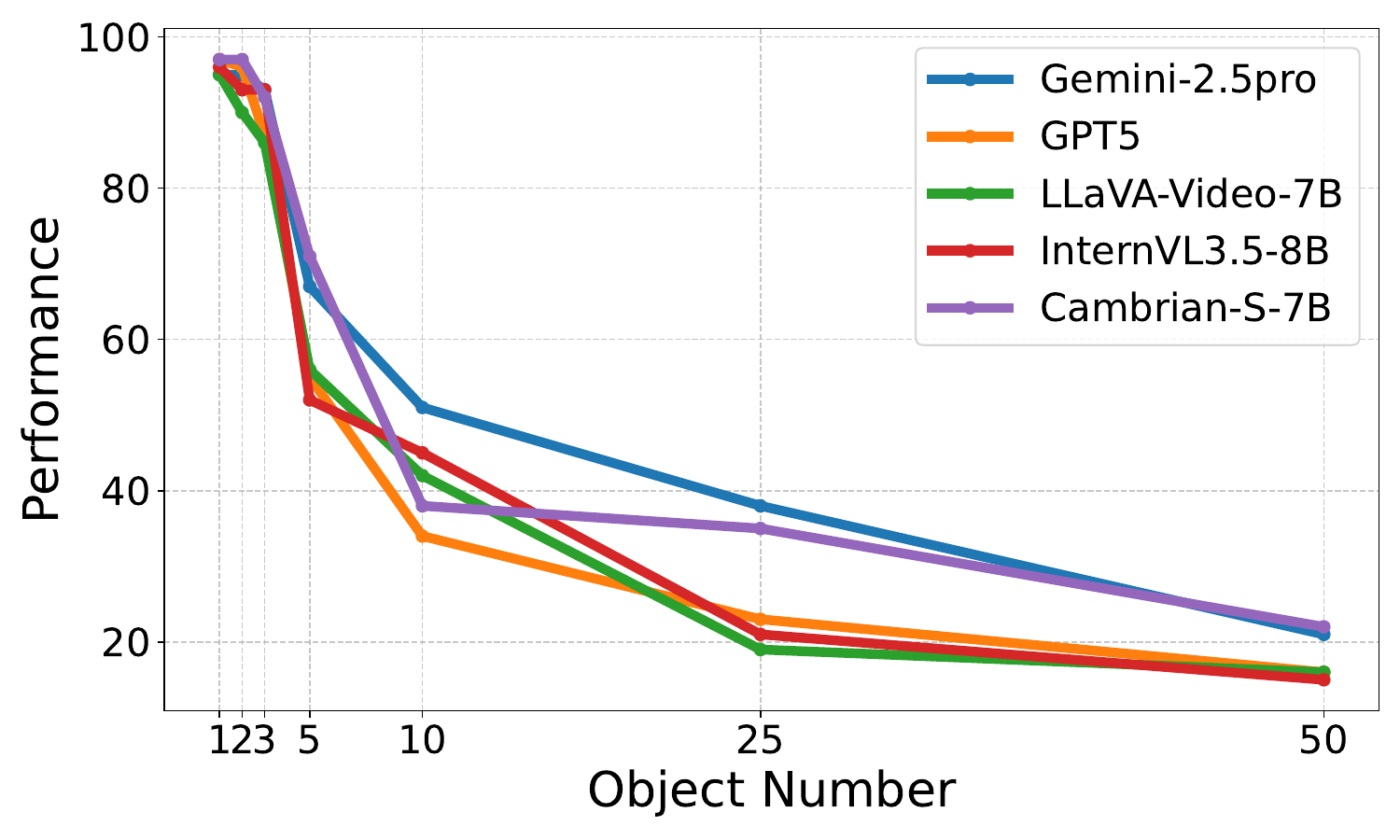}
	\vspace{-0.1in}
	\caption{VLMs' performance of spatial reasoning with varying compositional scene complexity}
\label{fig:object_count}
\end{figure}

We constructed benchmarks with the number of objects from 5 to 50, as shown in Figure \ref{fig:number}. Results in Figure~\ref{fig:object_count} showed that the reasoning performance of VLMs declines markedly with more objects, revealing a shared difficulty in managing more complicated visual clutter. Gemini-2.5-Pro and Cambrian-S-7B achieve higher accuracy compared to other models, as the scene complexity grows. We hypotheisze that the performance degradation with high compositional scene complexity stems from repetitive counting, since the same object could  appears multiple times in the input video, an effect reflected in the models' tendency to overestimate counts relative to the ground truth.

\begin{figure}[ht]
	\includegraphics[width=\linewidth]{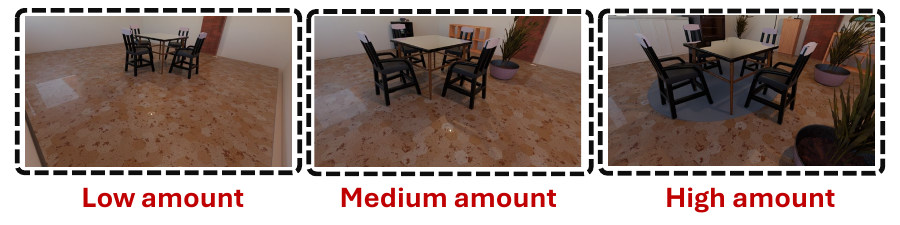}
	\vspace{-0.3in}
	\caption{Examples of benchmark samples with varying relational complexity (i.e., the amount of irrelevant objects)}
	\label{fig:irr}
	\vspace{-0.1in}
\end{figure}

\subsection{VLM's Spatial Reasoning with Varying Relational Scene Complexity}
We investigated how VLM's reasoning is influenced by irrelevant objects, which increase the complexity of spatial relations in the scene, as shown in Figure \ref{fig:irr}. 
Results in Table ~\ref{tab:combined_performance} show that increasing the number of irrelevant objects slightly reduces model performance. By examining the models’ reasoning paths, we found that this degradation is likely caused by incorrect object referencing.

\begin{figure}[ht]
	\includegraphics[width=\linewidth]{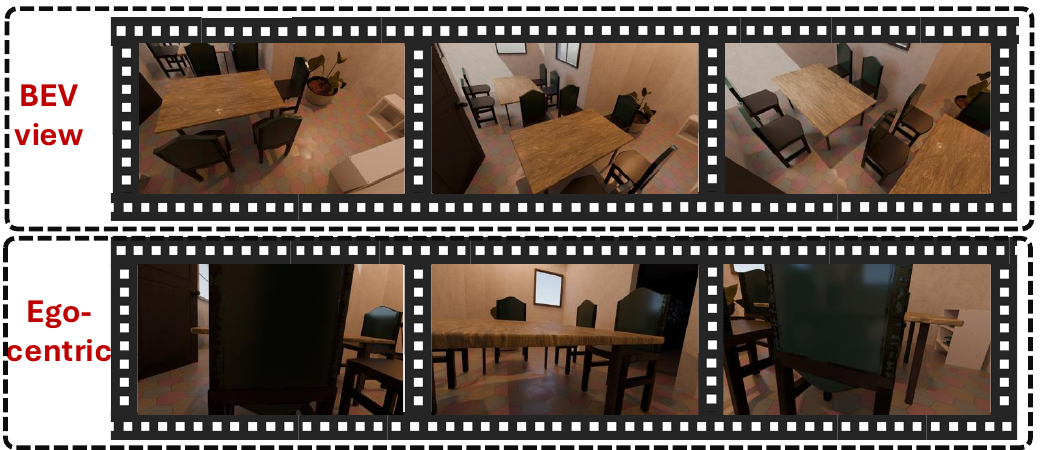}
	\caption{Examples of benchmark samples with varying observational complexity (i.e., different camera perspectives)}
	\label{fig:cam_height}
\end{figure}

\subsection{VLM's Spatial Reasoning with Varying Observational Scene Complexity}
We examined the impact of camera perspective on VLM performance. As shown in Figure \ref{fig:cam_height}, we generated two video variants for each scene: a bird’s-eye view captured from a higher camera position (2.5 m) and an egocentric view from a lower perspective (1 m). Results in Table~\ref{tab:observational} show that, for Perspective-taking and Spatiotemporal tasks, the model performance improves substantially as the viewpoint becomes closer to the bird’s-eye view. This enhancement may result from the fact that higher perspectives provide a clearer and more comprehensive view of the scene’s spatial layout. On the other hand, negligible differences are noticed on Measurement tasks.

\begin{table}[h!]
\centering
\small
{\fontsize{8}{9}\selectfont
\begin{tabular}{llccc}
	\cmidrule(lr){1-5} 
	& & \textbf{Measure} & \textbf{Perspect} & \textbf{Spatiotemporal} \\ 
	\midrule 
	\multirow{2}{*}{Gemini-2.5} & BEV & 68.1 & 77.4& 73.2 \\ 
	
	& Ego & 68.9 & 67.2& 70.1 \\ 
	\cmidrule(lr){2-5}
	\multirow{2}{*}{GPT5} & BEV & 42.0& 69.1& 49.0\\ 
	
	& Ego &41.2 & 55.5& 31.3 \\ 
	\cmidrule(lr){2-5} 
	\multirow{2}{*}{LLaVA-Video-7B} & BEV & 45.0& 68.3&47.1 \\ 
	
	& Ego &45.2 &56.1 & 36.2 \\ 
	\cmidrule(lr){2-5} 
	\multirow{2}{*}{InternVL3.5-8B} & BEV & 47.6& 62.7& 68.1\\ 
	
	& Ego &48.2 &52.9 &64.9 \\ 
	\cmidrule(lr){2-5} 
	\multirow{2}{*}{Cambrian-S-7B} & BEV& 57.5& 76.2&78.8 \\ 
	
	& Ego & 57.0& 71.3& 72.4\\ 
	\bottomrule 
\end{tabular}}
\caption{VLM's reasoning performance with varying observational scene complexity}
\label{tab:observational}
\end{table}

\section{Conclusion}
In this paper, we introduced InfiniBench, a fully automated benchmark generator that creates complex and physically plausible 3D scenes with customizable scene complexity. By enabling the systematic variation of scene properties, InfiniBench facilitates fine-grained and diagnostic analysis of VLM failures in spatial reasoning, paving the way for developing more robust and spatially-aware VLMs.

\bibliographystyle{abbrvnat}
\bibliography{main}

\appendix

\section{In-Context Learning for Scene Constraint Generation}
\label{sec:appendix_a}

In our design of InfiniBench, the LLM takes the following inputs: the user's scene description in natural language, the given descriptions about the list of available assets and constraint APIs (determined by the specific procedural generation framework being used). Besides, to ensure the generated scene constraints are machine-readable and physically plausible, in-context examples are also provided to the LLM to demonstrate the specific syntax of the underlying constraint language used in the procedural generation framework. In practice, we use the default home constraints which are manually crafted as the in-context examples. Some more specific examples, with respect to the constraint APIs and syntax being used in the Infinigen \cite{raistrick2024infinigen} framework, are provided below.

\subsection{Example 1: Dining Rooms}
The following code block demonstrates how to define scene constraints for a dining room, including furniture relationships (chairs around tables) and clutter distribution (dishware and utensils).

\begin{lstlisting}[language=Python, caption={An in-context example for Dining Room constraints provided to the LLM agent.}, label={lst:dining_example}]
	# region DININGROOMS
	
	diningtables = furniture[Semantics.Table][tables.TableDiningFactory]
	diningchairs = furniture[Semantics.Chair][seating.ChairFactory]
	constraints["dining_chairs"] = rooms.all(
	lambda r: (
	diningtables.related_to(r).all(
	lambda t: (
	diningchairs.related_to(r)
	.related_to(t, cu.front_against)
	.count()
	.in_range(3, 6)
	)
	)
	)
	)
	
	score_terms["dining_chairs"] = rooms.all(
	lambda r: (
	diningchairs.related_to(r).count().maximize(weight=5)
	+ diningchairs.related_to(r)
	.mean(lambda t: t.distance(diningchairs.related_to(r)))
	.maximize(weight=3)
	# cl.reflectional_asymmetry(diningchairs.related_to(r), diningtables.related_to(r)).minimize(weight=1)
	# cl.rotational_asymmetry(diningchairs.related_to(r)).minimize(weight=1)
	)
	)
	
	constraints["dining_table_objects"] = rooms.all(
	lambda r: (
	diningtables.related_to(r).all(
	lambda t: (
	obj[Semantics.TableDisplayItem]
	.related_to(t, cu.ontop)
	.count()
	.in_range(0, 2)
	* (obj[Semantics.Utensils].related_to(t, cu.ontop).count() >= 0)
	* (
	obj[Semantics.Dishware]
	.related_to(t, cu.ontop)
	.count()
	.in_range(0, 2)
	)
	)
	)
	)
	)
	
	score_terms["dining_table_objects"] = rooms.mean(
	lambda r: (
	cl.center_stable_surface_dist(
	obj[Semantics.TableDisplayItem].related_to(
	diningtables.related_to(r), cu.ontop
	)
	).minimize(weight=1)
	)
	)
	
	diningrooms = rooms[Semantics.DiningRoom].excludes(cu.room_types)
	constraints["diningroom"] = diningrooms.all(
	lambda r: (
	(diningtables.related_to(r).count() == 1)
	* storage.related_to(r).all(
	lambda t: (
	(obj[Semantics.Dishware].related_to(t, cu.on).count() >= 0)
	* (
	obj[Semantics.OfficeShelfItem]
	.related_to(t, cu.on)
	.count()
	.in_range(0, 5)
	)
	)
	)
	)
	)
	score_terms["diningroom"] = diningrooms.mean(
	lambda r: (
	diningtables.related_to(r).distance(r, cu.walltags).maximize(weight=10)
	+ cl.angle_alignment_cost(
	diningtables.related_to(r), r, cu.walltags
	).minimize(weight=10)
	+ cl.center_stable_surface_dist(diningtables.related_to(r)).minimize(
	weight=1
	)
	)
	)
	# endregion
\end{lstlisting}

\subsection{Example 2: Bathrooms}
The following example illustrates the logical constraints for bathroom layouts, specifically enforcing the presence of essential hardware (toilets, sinks) and their spatial accessibility.

\begin{lstlisting}[language=Python, caption={An in-context example for Bathroom constraints provided to the LLM agent.}, label={lst:bathroom_example}]
	# region BATHROOMS
	bathrooms = rooms[Semantics.Bathroom].excludes(cu.room_types)
	
	toilet = wallfurn[bathroom.ToiletFactory]
	bathtub = wallfurn[bathroom.BathtubFactory]
	sink = wallfurn[bathroom.StandingSinkFactory]
	
	hardware = obj[bathroom.HardwareFactory].related_to(bathrooms, cu.against_wall)
	
	constraints["bathroom"] = bathrooms.all(
	lambda r: (
	mirror.related_to(r).related_to(r, cu.flush_wall).count().equals(1)
	* sink.related_to(r).count().equals(1)
	* toilet.related_to(r).count().equals(1)
	* storage.related_to(r).all(
	lambda t: (
	obj[Semantics.BathroomItem].related_to(t, cu.on).count() >= 0
	)
	)
	)
	)
	
	score_terms["toilet"] = rooms.all(
	lambda r: (
	toilet.distance(doors).maximize(weight=1)
	+ toilet.distance(furniture).maximize(weight=1)
	+ toilet.distance(sink).maximize(weight=1)
	+ cl.accessibility_cost(toilet, furniture, dist=2).minimize(weight=10)
	)
	)
	
	constraints["bathtub"] = bathrooms.all(
	lambda r: (
	bathtub.related_to(r).count().in_range(0, 1)
	* hardware.related_to(r).count().in_range(1, 4)
	)
	)
	score_terms["bathtub"] = bathrooms.all(
	lambda r: (
	bathtub.mean(lambda t: t.distance(hardware)).minimize(weight=0.2)
	+ sink.mean(lambda t: t.distance(hardware)).minimize(weight=0.2)
	+ hardware.mean(
	lambda t: (
	t.distance(rooms, cu.floortags).hinge(0.5, 1).minimize(weight=15)
	)
	)
	)
	)
	
	score_terms["bathroom"] = (
	mirror.related_to(bathrooms).distance(sink).minimize(weight=3)
	) + cl.accessibility_cost(mirror, furniture, cu.down_dir).maximize(weight=3)
	# endregion
\end{lstlisting}

\section{CoT Reasoning for Refinement of Scene Constraints}
\label{sec:appendix_b}

This section details the iterative feedback loop employed to resolve the conflicts in the generation of scene constraints. The refinement process operates as follows:

\begin{enumerate}
	\item \textbf{Execution:} The layout optimizer attempts to generate the scene based on the current set of procedural constraints.
	
	\item \textbf{Error Reporting:} Upon the completion of a fixed number of optimization steps, a structured error report is synthesized from the optimization metadata. This report comprises:
	\begin{itemize}
		\item \textit{Textual Summary:} A list of constraints that remained unsatisfied during the optimization process.
		\item \textit{Visual Metadata:} A semantic 2D BEV (Bird's-Eye View) segmentation map annotated with object labels to visualize spatial conflicts.
	\end{itemize}
	
	\item \textbf{Refinement:} Based on the user's original scene description and the generated error report, the LLM evaluates whether the user's requirements have been fully met. If unsatisfied, the LLM employs Chain-of-Thought (CoT) reasoning to analyze the failure and output revised constraints. In practice, we use the thinking mode of Gemini-2.5-Pro to enable CoT reasoning.
\end{enumerate}

A detailed end-to-end example of such CoT reasoning, with the constraint APIs and syntax provided by the Infinigen framework \cite{raistrick2024infinigen},  is provided in Figure \ref{fig:refinement_eg} below: 

\begin{figure}[ht]
	\centering
	\includegraphics[width=\linewidth]{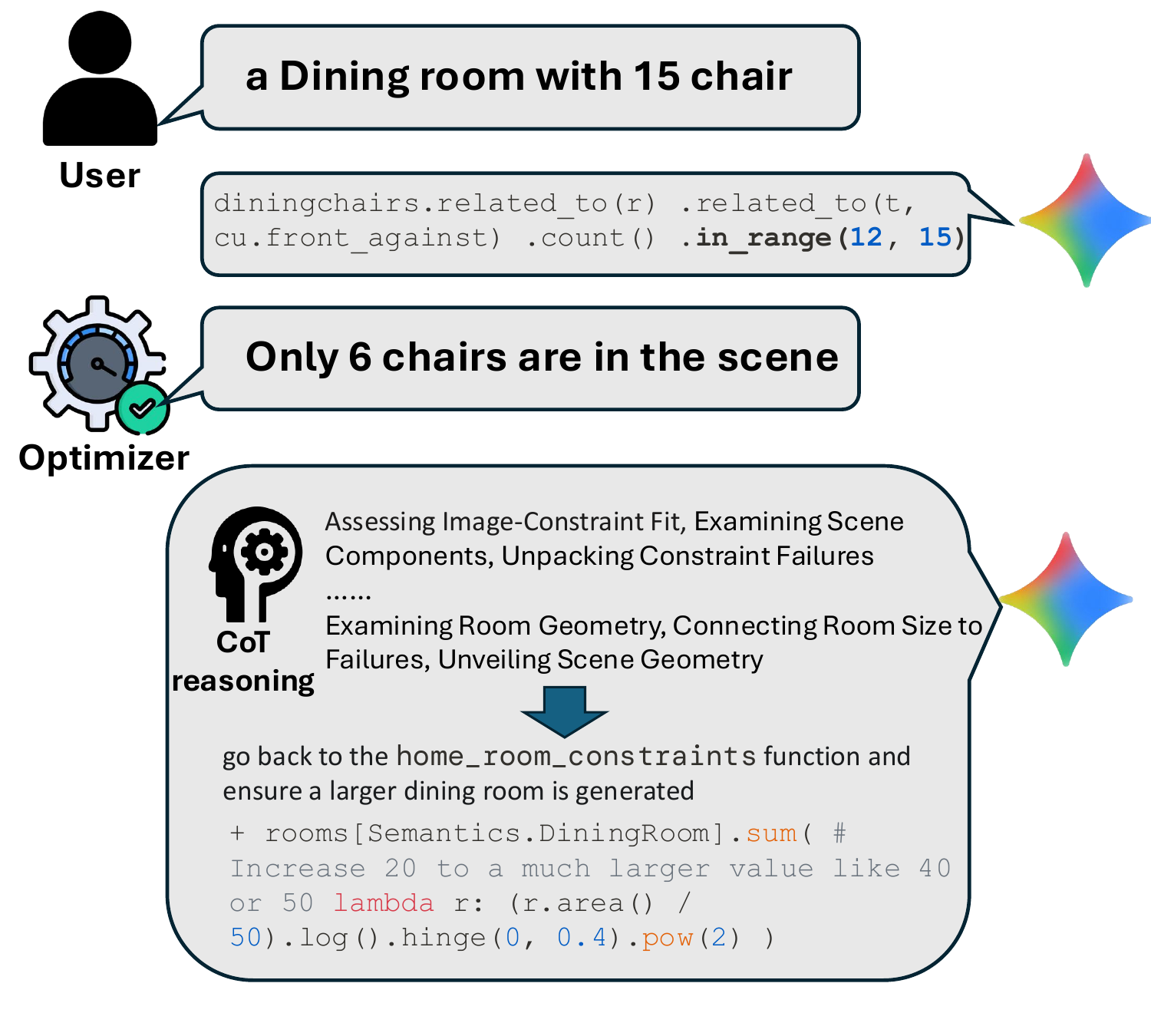}
	\caption{An example of scene constraint refinement via CoT reasoning}
	\label{fig:refinement_eg}
\end{figure}

\begin{figure*}[h]
	\centering
	\includegraphics[width=\linewidth]{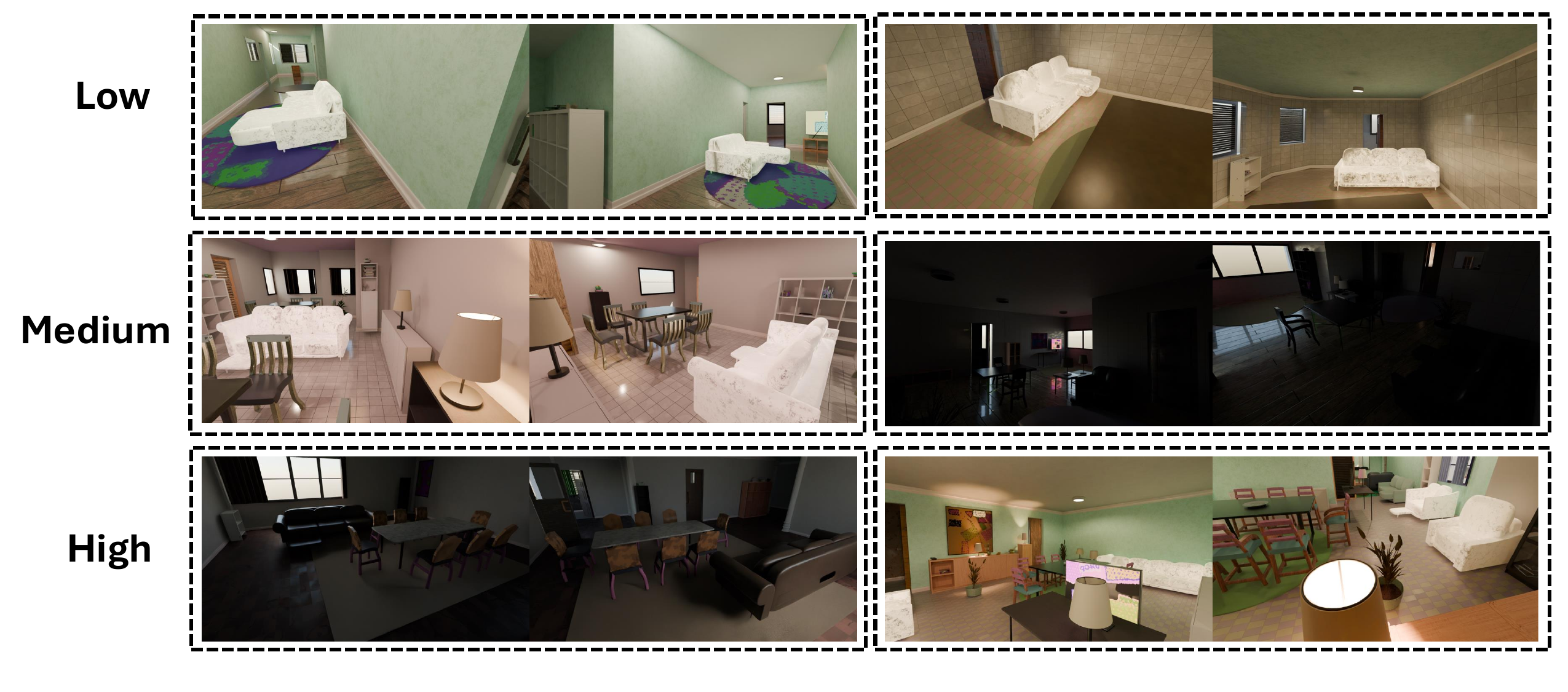}
	\caption{Examples of generated \textbf{living rooms} with varying levels of \textbf{compositional complexity}, measured by the number of objects. `Low' indicates seating for fewer than five people; `Medium' indicates seating for five to ten people; and `High' indicates seating for more than ten people. The input prompt to the LLM agent for scene generation is: ``Generate a living room with seating for \textit{$<$num$>$} people''.}
	\label{fig:example_compositional_living}
\end{figure*}

\begin{figure*}[h]
	\centering
	\includegraphics[width=\linewidth]{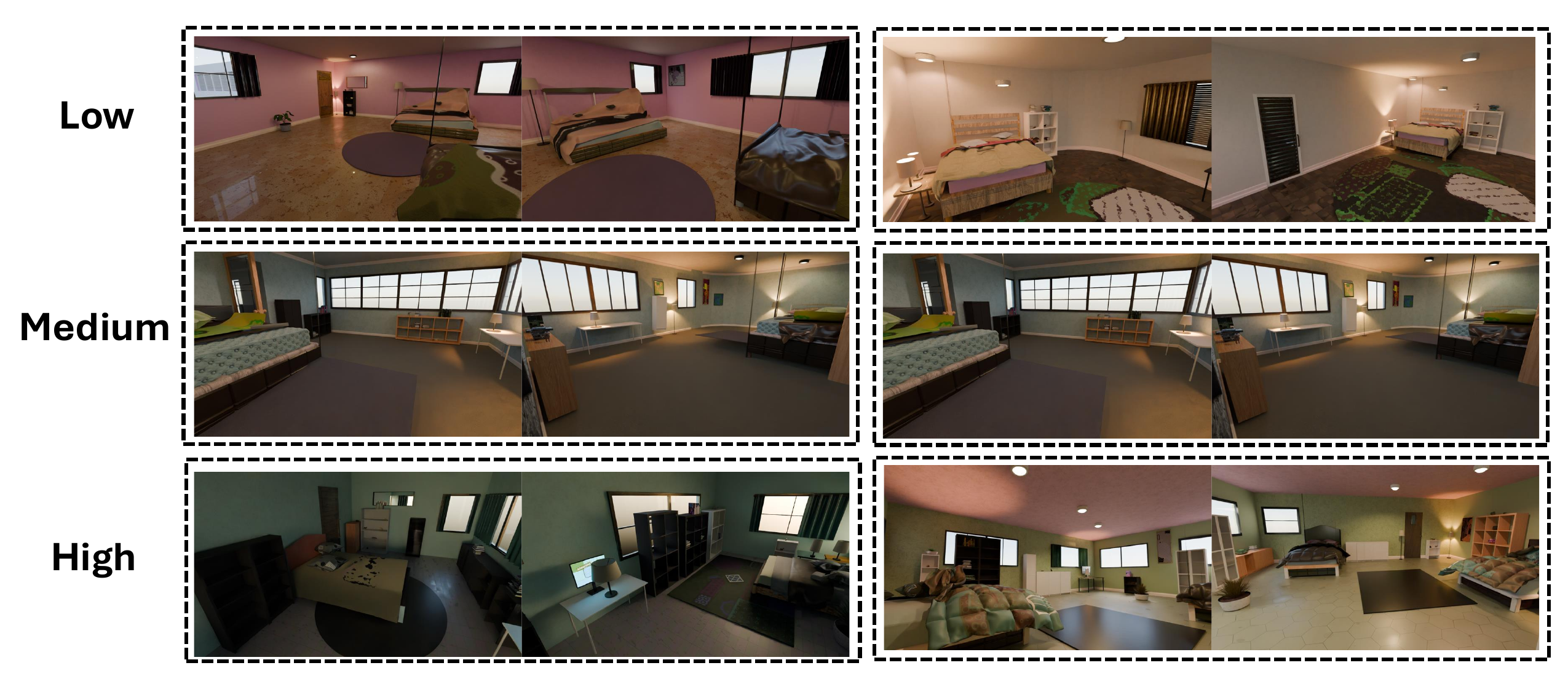}
	\caption{Examples of generated \textbf{bedrooms} with varying levels of \textbf{compositional complexity}, measured by the number of objects. `Low' indicates only one shelf; `Medium' indicates 2-8 shelves; and `High' indicates seating for more than 10 shelves.  The input prompt to the LLM agent for scene generation is: ``Generate a bedroom with \textit{$<$num$>$} shelves''.}
	\label{fig:example_compositional_bedroom}
\end{figure*}

\begin{figure*}[h]
	\centering
	\includegraphics[width=\linewidth]{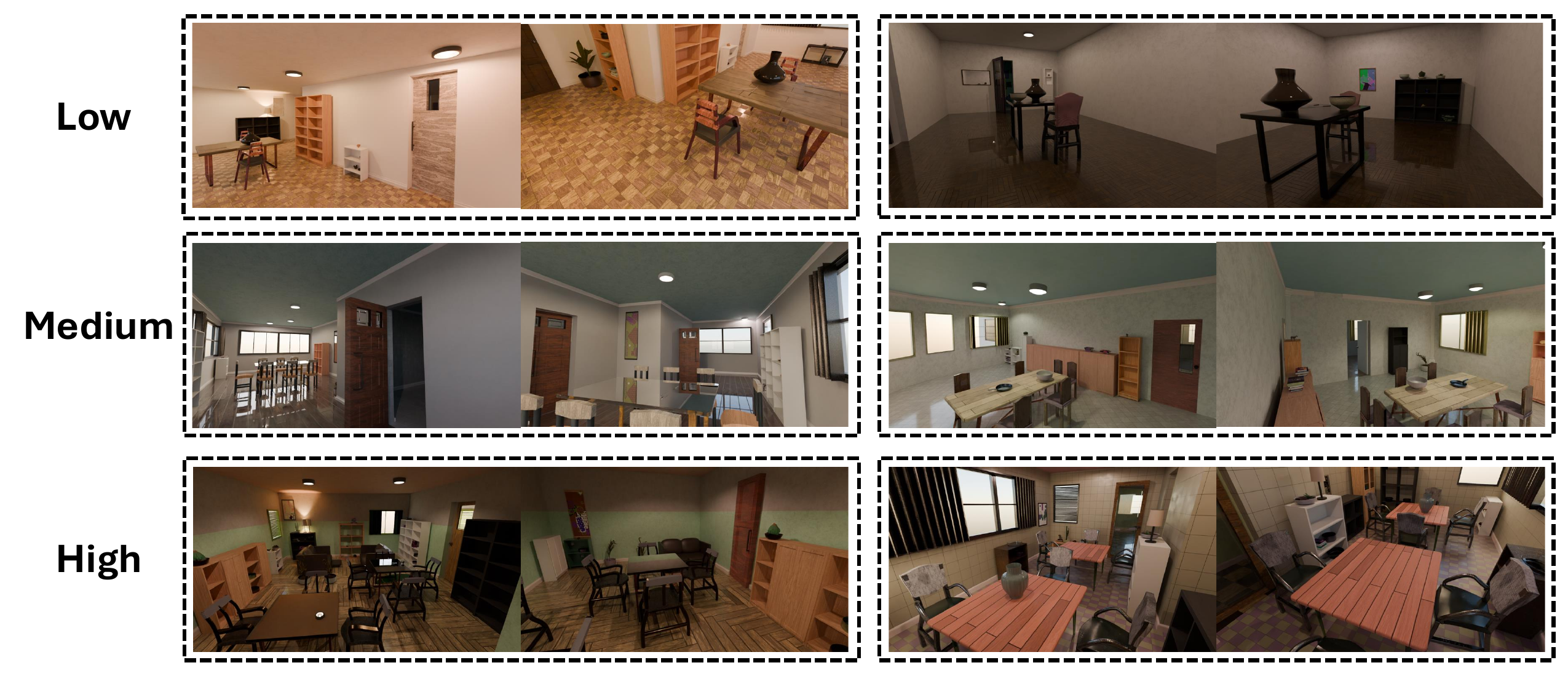}
	\caption{Examples of generated \textbf{dining rooms} with varying levels of \textbf{compositional complexity}, measured by the number of objects. `Low' indicates seating for fewer than five people; `Medium' indicates seating for five to ten people; and `High' indicates seating for more than ten people. The input prompt to the LLM agent for scene generation is: ``Generate a dining room with seating for \textit{$<$num$>$} people''.}
	\label{fig:example_compositional_dinning}
\end{figure*}

\section{Details of Cluster-Based Optimization}
\label{sec:appendix_c}

To address the limitations of the ``rigid hierarchy'' that is prevalent in traditional procedural generation frameworks, in InfiniBench we implemented a robust cluster-based optimization strategy.

\subsection{Cluster Definition and Identification}
We define the \textbf{root} of a cluster as an object situated at the highest level of the hierarchy (typically large furniture, such as tables or desks). Objects at lower hierarchy levels are designated as \textbf{child} objects if they maintain a ``stable against'' spatial relationship with a parent object. Information regarding these clusters is dynamically retrieved and updated from the global ``scene state'' throughout the optimization process.

\subsection{Augmented Action Space}
In standard rigid-hierarchy optimization, the action space for individual objects comprises addition, resampling, deletion, translation, rotation, relation plane changes, relation target changes, and swapping. To facilitate cluster-based optimization, we augment this space by adding the following cluster-specific operations:

\begin{itemize}
	\item \texttt{resample\_cluster}: Resamples the parameters for the cluster's root object, triggering a cascade update for child objects.
	\item \texttt{translation\_cluster}: Translates the entire cluster as a single entity while strictly preserving the relative positions of all constituent objects.
	\item \texttt{rotation\_cluster}: Rotates the cluster around the geometric center of the root object, maintaining the relative orientation and position of all child objects.
\end{itemize}

\subsection{Collision Checking}
Following each operation, the optimizer performs a collision detection and checking routine. If a collision is detected, the operation is immediately reverted to maintain physical plausibility. For cluster-level operations, collision checks are performed using the cluster's collective bounding box (or the aggregate of all component meshes) to ensure the entire group is placed validly without intersecting other scene elements.

\section{Gallery of Generated 3D Scenes}
\label{sec:appendix_d}

\begin{figure*}[h]
	\centering
	\includegraphics[width=\linewidth]{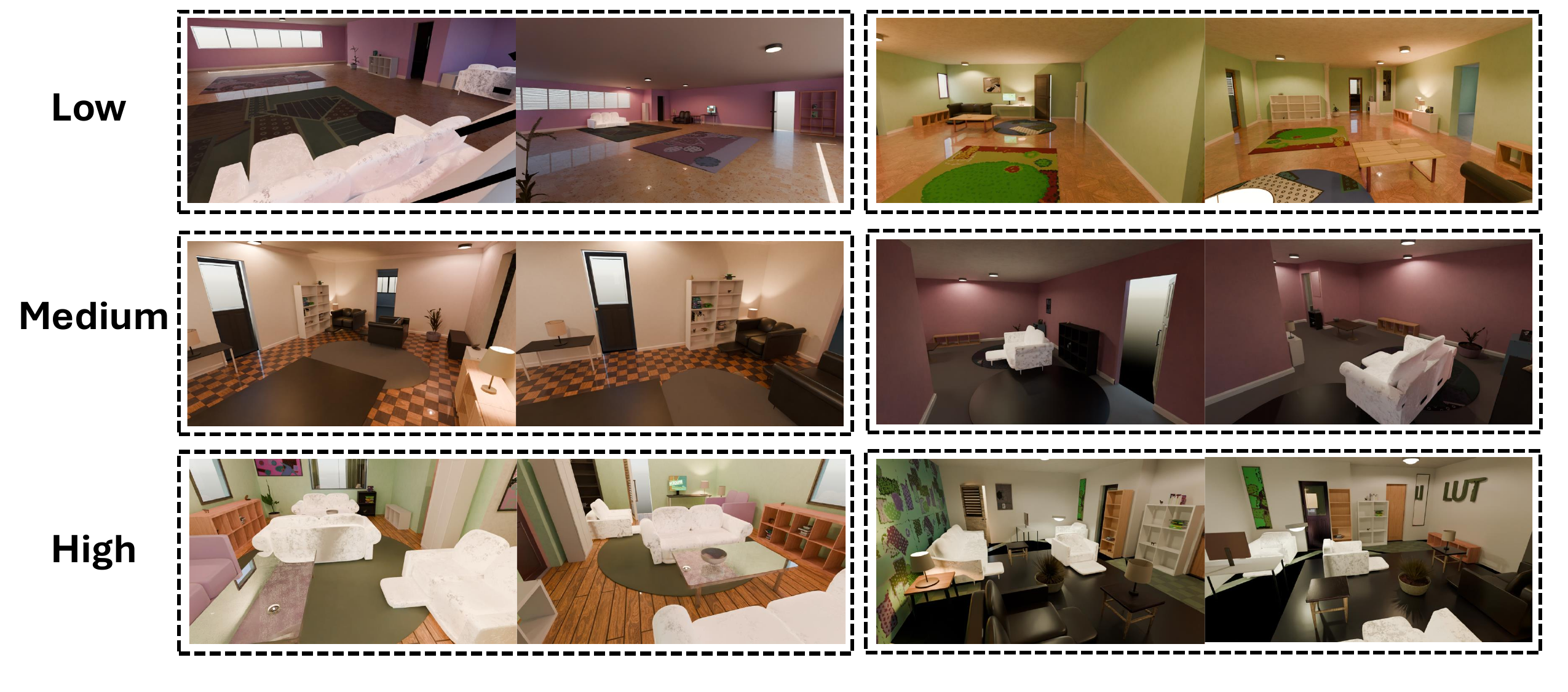}
	\caption{Examples of generated \textbf{living rooms} with varying levels of \textbf{relational complexity}, measured by the room occupancy ratio. `Low' indicates an occupancy ratio below 10\%; `Medium' indicates an occupancy ratio between 10\% and 50\%; and `High' indicates an occupancy rate above 50\%. The input prompt to the LLM agent for scene generation is: ``Generate a living room with a room occupancy ratio of \textit{$<$ratio$>$}''.}
	\label{fig:example_relational_living}
\end{figure*}

\begin{figure*}[h]
	\centering
	\includegraphics[width=\linewidth]{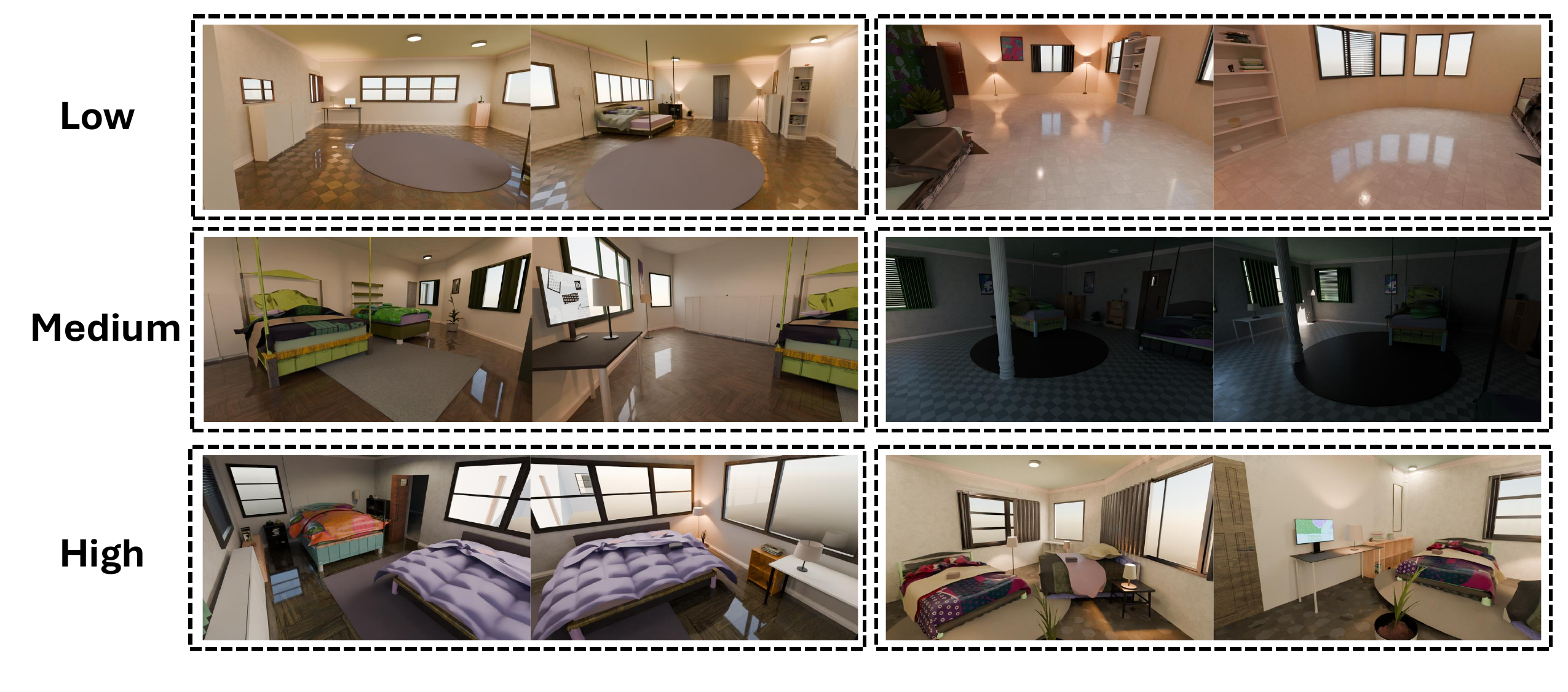}
	\caption{Examples of generated \textbf{bedrooms} with varying levels of \textbf{relational complexity}, measured by the room occupancy ratio.  `Low' indicates an occupancy ratio below 10\%; `Medium' indicates an occupancy ratio between 10\% and 50\%; and `High' indicates an occupancy rate above 50\%. The input prompt to the LLM agent for scene generation is: ``Generate a bedroom with a room occupancy ratio of \textit{$<$ratio$>$}''.}
	\label{fig:example_relational_bedrom}
\end{figure*}

\begin{figure*}[h]
	\centering
	\includegraphics[width=\linewidth]{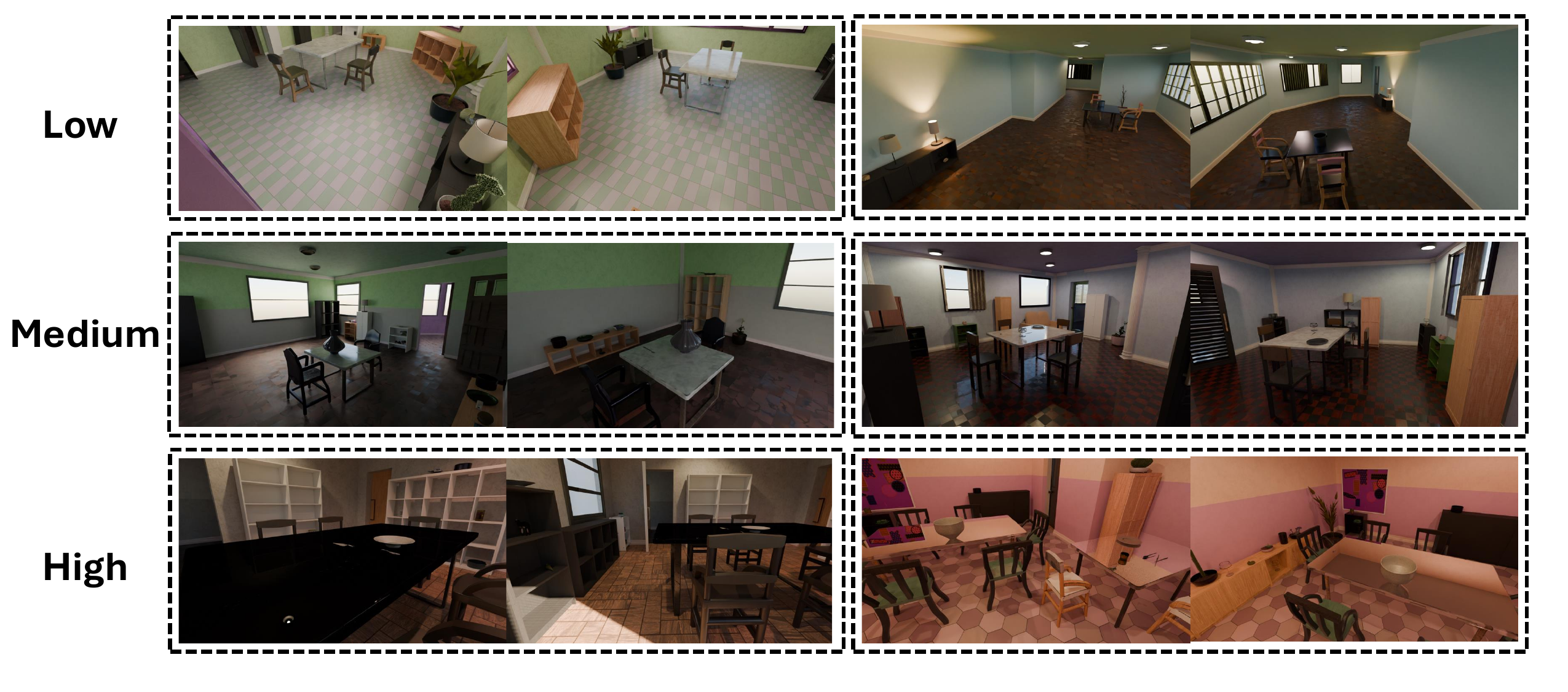}
	\caption{Examples of generated \textbf{dining rooms} with varying levels of \textbf{relational complexity}, measured by the room occupancy ratio. `Low' indicates an occupancy ratio below 10\%; `Medium' indicates an occupancy ratio between 10\% and 50\%; and `High' indicates an occupancy rate above 50\%. The input prompt to the LLM agent for scene generation is: ``Generate a dining room with a room occupancy ratio of \textit{$<$ratio$>$}''.}
	\label{fig:example_relational_dinning}
\end{figure*}

\begin{figure*}[h]
	\centering
	\includegraphics[width=0.7\linewidth]{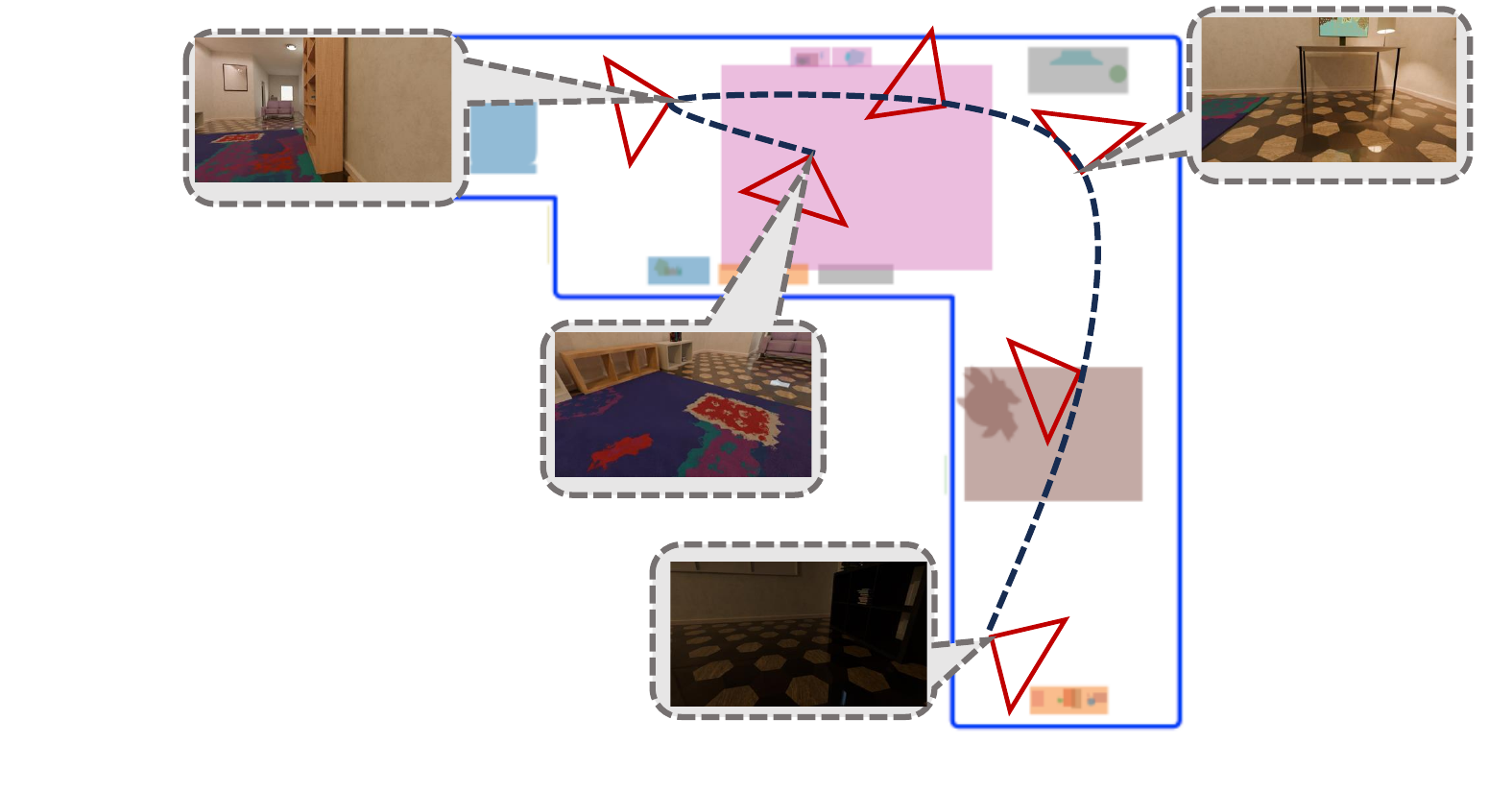}
	\caption{Visualization of the camera trajectory under a specific level of observational complexity (partial occlusion). The input prompt to the LLM agent for scene generation is: ``Generate a typical living room. The camera should capture objects with partial occlusion''.}
	\label{fig:cam_occ}
\end{figure*}

\begin{figure*}[h]
	\centering
	\includegraphics[width=0.7\linewidth]{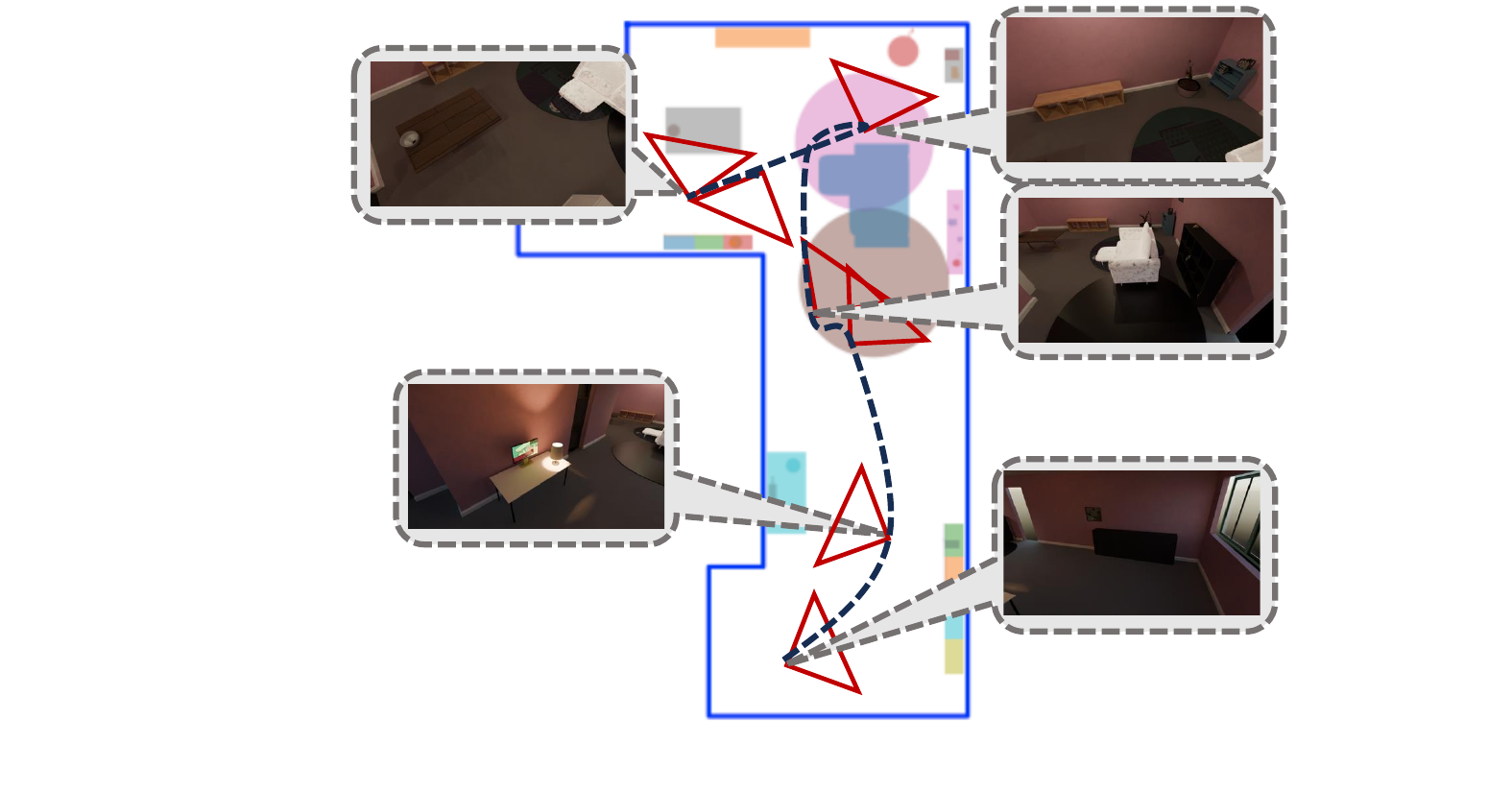}
	\caption{Visualization of the camera path under a specific level of observational complexity (bird-eye view, camera height is 2m). The input prompt to the LLM agent for scene generation is: ``Generate a typical living room. The camera's capture height should be 2 meters''.}
	\label{fig:cam_bev}
\end{figure*}

We provide additional examples of the 3D scenes generated by InfiniBench, demonstrating its versatility across various room types and levels of scene complexity. The scenes with different levels of compositional complexity are shown in Figures \ref{fig:example_compositional_living}-\ref{fig:example_compositional_dinning}, and the scenes with different levels of relational complexity are shown in Figures \ref{fig:example_relational_living}-\ref{fig:example_relational_dinning}. For each generated scene, we showed images captured from two viewpoints, for better visualization of the generated room layouts and furniture positioning. In addition, in Figure \ref{fig:cam_occ} and Figure \ref{fig:cam_bev}, we also provided examples of the optimized camera trajectories, along with the scene frames at the selected waypoints, with different levels of observational complexities. Note that, in addition to InfiniBench’s native scene-generation pipeline, we also leverage the room-shape generation module from Infinigen to produce more realistic renderings of these example scenes.

\section{Details of the Camera Trajectory Optimization Algorithm}
\label{sec:appendix_e}

To ensure that the generated camera trajectory simulates a natural, human-like perspective while maximizing the visual coverage of task-related objects, we constrain the optimization problem to a 4-DOF space. We fix the camera's vertical height ($z$) to approximate a standard eye level and restrict the roll angle ($\theta_x$) to zero to maintain view stability. Consequently, the optimization is performed over the remaining four degrees of freedom: the horizontal position coordinates $(x, y)$ and the rotational angles for yaw ($\theta_y$) and pitch ($\theta_z$). This configuration allows the camera to freely navigate accessible areas on the horizontal plane and adjust its viewing angle to effectively frame relevant spatial features.

\noindent\textbf{Optimization Workflow.}
As shown in Algorithm \ref{algo:path_algo}, our method follows an iterative workflow that progressively builds a path to view every task-related object in the scene. The process is as follows:

\noindent\textit{0. Scene Preprocessing:}
Before optimization, the generated scene file (e.g., a \texttt{.blend} file) is preprocessed to extract the necessary geometric and semantic metadata. Using Blender's \texttt{Bmesh} API, we extract the 3D mesh of every object. These meshes are then triangulated and converted into \texttt{Trimesh} objects, which facilitates efficient geometric queries. A 2D floor plan is also extracted to define the navigable area for the camera.

\noindent\textit{1. Target Selection (find the closest target object):}
The algorithm begins at a predefined starting point, typically a door. It maintains a list of ``unvisited'' objects. In each iteration, it identifies the object from this list that is closest to the camera's current position, designating it as the next target.

\noindent\textit{2. Viewpoint sampling:}
A set of candidate viewpoints is randomly sampled in the vicinity of the current target object. Each viewpoint is a 4-DOF pose that is oriented towards the object.

\noindent\textit{3. Viewpoint selection:}
Each candidate viewpoint is evaluated against a series of physical constraints to determine its validity. The best viewpoint is one that satisfies all the following criteria:
\begin{itemize}
	\itemsep0em 
	\item \emph{Accessibility}: The camera's \((x,y)\) location must be within the accessible area of the 2D floor plan (i.e., inside the room but outside the bounding boxes of all objects).
	\item \emph{Field of View (FOV)}: The target object must be fully visible within the camera's viewing frustum from the candidate pose.
	\item \emph{Occlusion Check}: To ensure a clear line of sight, we perform an occlusion check. A segmentation map is rendered from the viewpoint using a fast off-screen renderer (\texttt{pyrenderer}). By projecting the 3D mesh of the target object and comparing it against the segmentation map's depth buffer, we calculate the percentage of the object that is occluded by other geometry. A viewpoint is only considered valid if the occlusion is below a predefined threshold.
\end{itemize}
The candidate that best satisfies these conditions (e.g., has the lowest occlusion) is selected as the next keyframe in the trajectory.

\noindent\textit{4. Path planning:}
Once a valid viewpoint is selected, we use Dijkstra's algorithm to compute the shortest, collision-free path on the 2D floor plan from the camera's current position to the new target viewpoint.

\noindent\textit{5. Iteration:}
The camera is moved along the planned path to the new viewpoint, the target object is marked as "visited," and the process repeats from Step 1. This loop continues until all objects in the scene have been successfully captured by at least one viewpoint. The final output is a complete, smooth camera trajectory composed of a sequence of optimal viewpoints connected by navigable paths, ensuring a comprehensive visual record of the scene.

\begin{algorithm}
	
	\caption{Camera Trajectory Optimization}
	\label{algo:path_algo}
	\begin{algorithmic}[1]
		\Require Scene, MaxSamplingTimes, MaxDistance, Threshold, RequiredRange
		
		\State $\textit{CamParams} \gets \Call{GetCameraParameters}{}$
		\State $\textit{Objects} \gets \Call{ExtractObjectMeshes}{\textit{Scene}}$
		\State $\textit{Targets} \gets \Call{GetCaptureList}{}$
		\State $\textit{CurrentView} \gets \Call{GetDoorPosition}{}$
		\State $\textit{TotalPath} \gets [\,]$
		\State $\textit{RemainingTargets} \gets \Call{Copy}{\textit{Targets}}$ 
		
		\While{$\neg\Call{IsEmpty}{\textit{RemainingTargets}}$} 
		\State $\textit{target} \gets \Call{FindClosestTarget}{\textit{CurrentView}}$\Comment{1. Target selection}
		\For{$i = 1$ to $\textit{MaxSamplingTimes}$}
		
		\State $\textit{view} \gets \Call{SampleViewpoint}{\textit{target}, \textit{MaxDistance}}$\Comment{2. Viewpoint sampling}
		\If{$\neg\Call{CheckCollision}{\textit{view}, \textit{Targets}}$}\Comment{3. Viewpoint selection}
		\State $\textit{FOV} \gets \Call{ComputeFOV}{\textit{view}, \textit{CamParams}}$
		\State $\textit{FOVPercent} \gets \Call{CheckFOV}{\textit{FOV}, \textit{target}}$
		\If{$\textit{FOVPercent} > \textit{Thr}$} 
		\State $\textit{SegMap} \gets \Call{Renderer}{\textit{view}, \textit{Scene}}$
		\State $\textit{Occ} \gets \Call{ComputeOcclusionRate}{\textit{SegMap}, \textit{target}}$
		\If{$\textit{Occ} \in \textit{RequiredRange}$}
		\State $\textit{Path} \gets \Call{PlanPathDijkstra}{\textit{CurrentView}, \textit{view},\textit{scene}}$ \Comment{4. Path planning}
		\State $\Call{Add}{\textit{TotalPath}, \textit{Path}}$
		\State $\textit{CurrentView} \gets \textit{view}$
		\State \textbf{break} 
		\EndIf
		\EndIf
		\EndIf
		\EndFor
		\State $\Call{Remove}{\textit{RemainingTargets}, \textit{target}}$ \Comment{5. Proceed to the next closest target}
		\EndWhile
		\State \Return $\textit{TotalPath}$
	\end{algorithmic}
\end{algorithm}

\begin{figure}[h]
	\centering
	\includegraphics[width=1.05\linewidth]{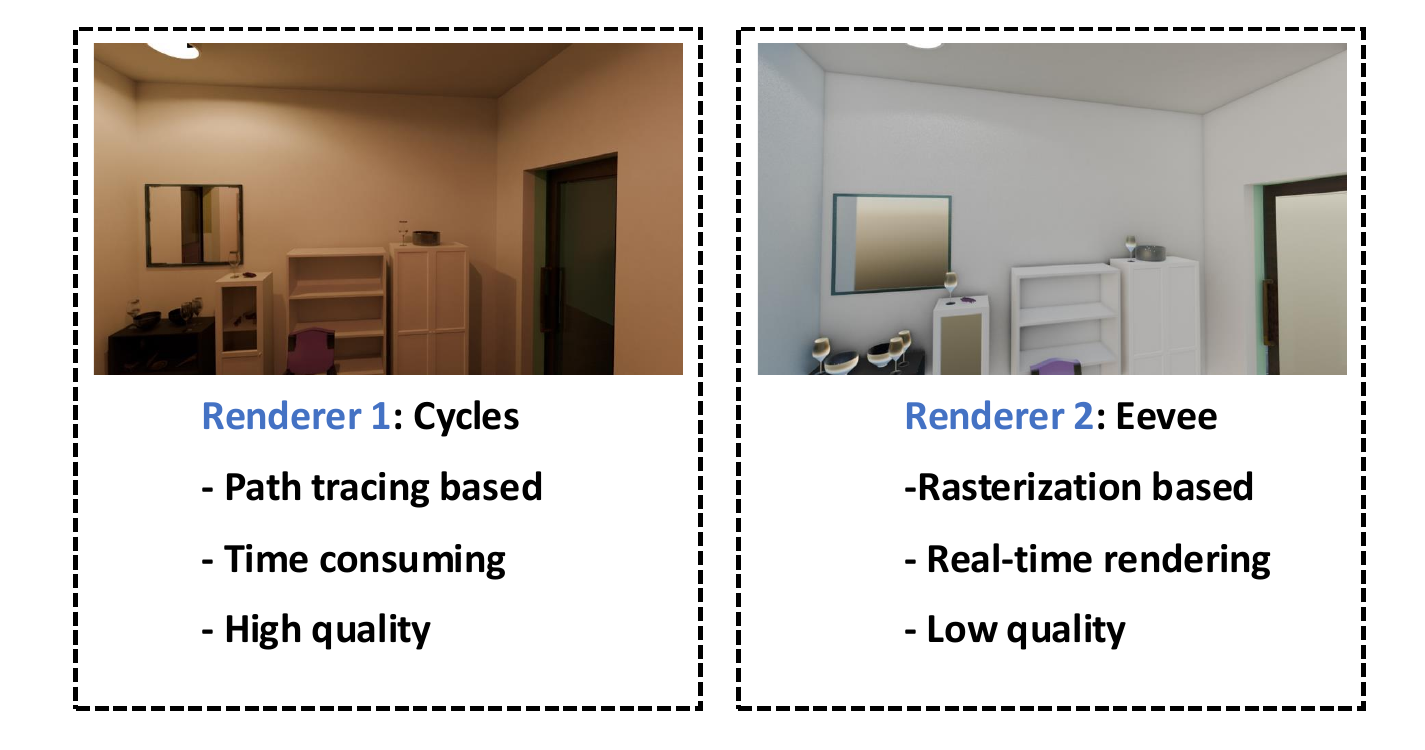}
	\caption{Comparison of image quality across different renderers.}
	\label{fig:renderer}
\end{figure}

\section{Details of InfiniBench Implementation}
\label{sec:appendix_f}
\begin{figure}[h]
	\centering
	\includegraphics[width=1.05\linewidth]{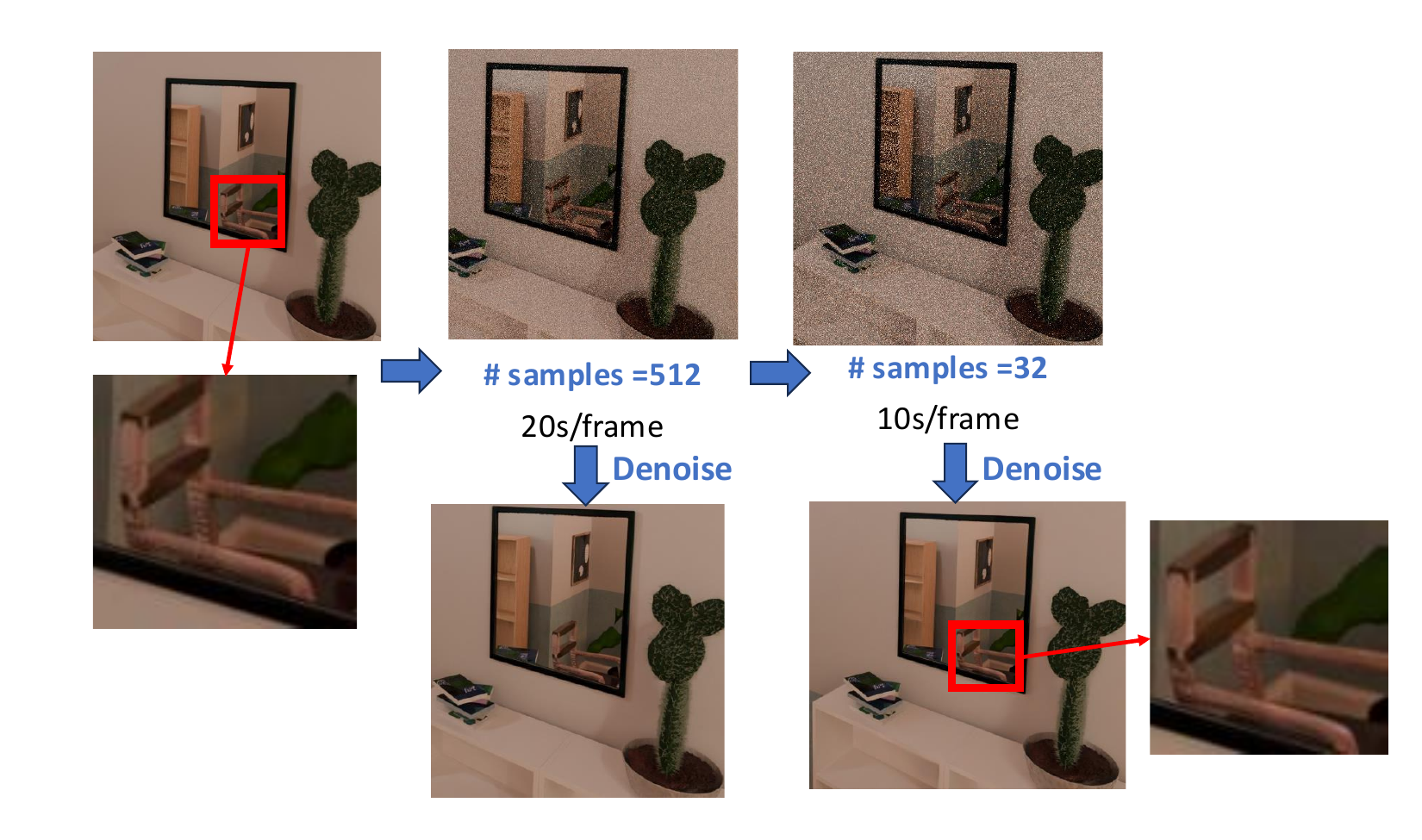}
	\caption{Impact of sample counts on image quality.}
	\label{fig:num_samples}
\end{figure}

As illustrated in Figure \ref{fig:renderer}, we employ the Blender Cycles engine, which utilizes path tracing, to achieve photorealistic frame rendering. A critical parameter in this process is the number of path-tracing samples. While 8192 samples are typically recommended to produce noise-free images (as shown in Figure \ref{fig:num_samples}), this setting is computationally expensive, requiring approximately 5 minutes per frame. To optimize for efficiency without compromising quality, we adjust the sample count to 32 and enable denoising. As shown in Figure \ref{fig:renderer}, after reducing the number of samples and enabling such denoising, the rendered frames appear nearly identical to the originals with 8192 samples, with only some fine textures slightly blurred. The specific rendering configuration is detailed below (unlisted parameters remain at default values).

\begin{itemize}
	\item \texttt{num\_samples}: 128
	\item \texttt{adaptive\_threshold}: 0.01
	\item \texttt{denoise}: True
	\item \texttt{exposure}: 1
\end{itemize}

\section{Additional Qualitative Comparisons}
\label{sec:appendix_g}
\begin{figure*}[h]
	\centering
	\includegraphics[width=\linewidth]{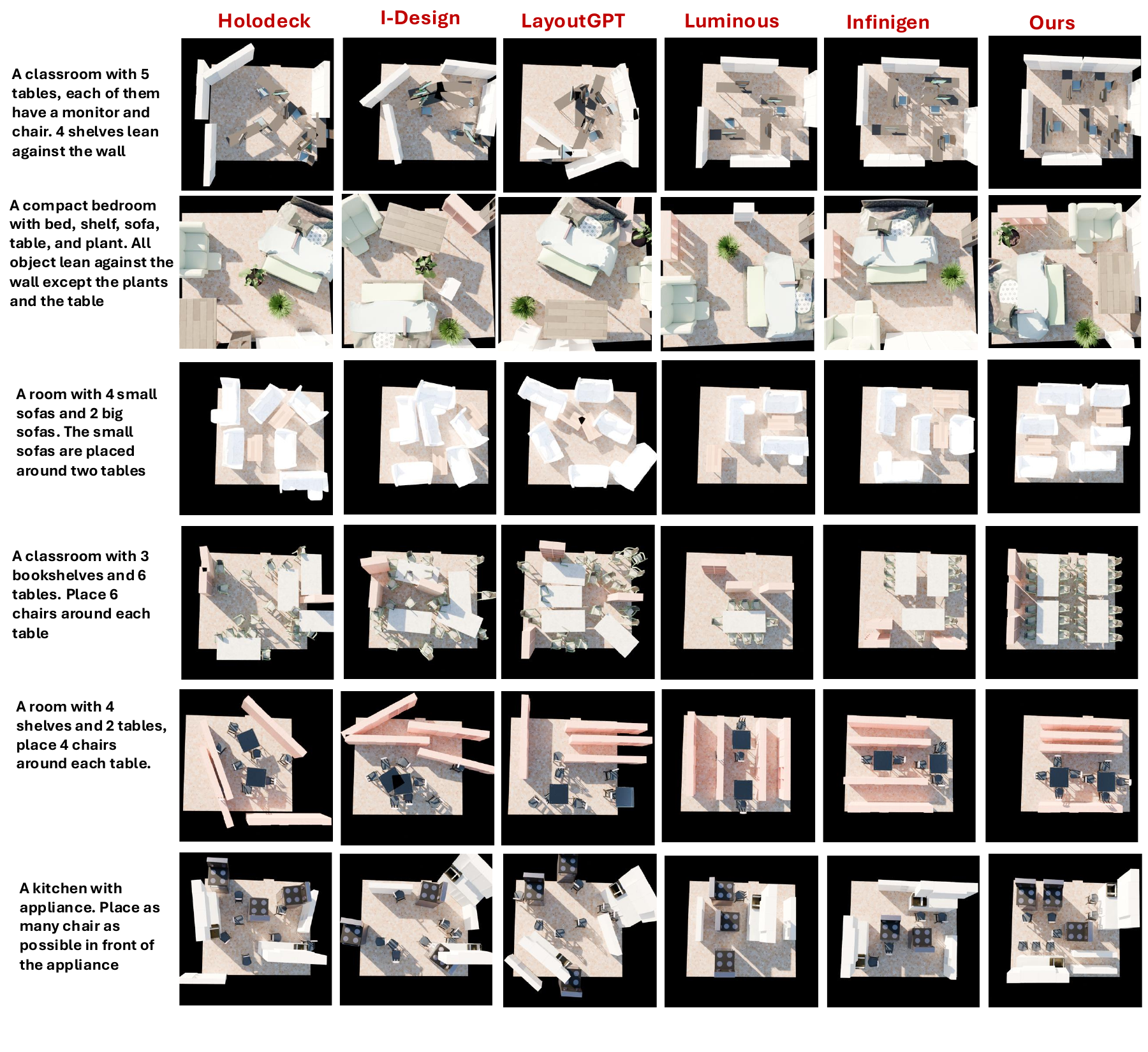}
	\caption{Comparing InfiniBench with baselines}
	\label{fig:more_comparisons}
\end{figure*}

As shown in Figure \ref{fig:more_comparisons}, we present additional examples comparing scenes generated by InfiniBench against baseline approaches. To ensure a fair comparison, all methods utilize the same 3D asset library and an identical square-shaped room plan. For baselines implemented in disparate engines, we exported the scene layouts and rendered them using Blender with identical settings to maintain visual consistency. For baselines incapable of processing natural language input directly, we employed a constraint generation method similar to Infinigen, iteratively refining the constraints until the program became executable.

\end{document}